\documentclass[journal]{IEEEtran}
\usepackage{amsthm}
\usepackage{cite}
 \usepackage[switch]{lineno}
\usepackage{url}
\usepackage{enumerate}
\newcommand{\XG}[1]{\textcolor[rgb]{0.00,0.00,1.00}{#1}}

\usepackage{graphicx,amsfonts,algorithmic,multirow,booktabs,color}
\usepackage[caption=false,font=footnotesize]{subfig}
\usepackage[linesnumbered,ruled,vlined]{algorithm2e}
\usepackage{amsmath,amssymb,multirow,soul,threeparttable,mathrsfs}

\definecolor{hl}{rgb}{0.75,0.75,0.75}
\definecolor{ColorName}{rgb}{0,0,1}
\sethlcolor{hl}

\begin{document}

%
\title{Accelerating Evolutionary Neural Architecture Search via Multi-Fidelity Evaluation
\thanks{Manuscript received --. This work was supported in part by the National Key Research and Development Project under Grant 2018AAA0100105, in part by the National Natural Science Foundation of China under Grant 61822301, 61876123, 61906001, and U1804262, in part by the Hong Kong Scholars Program under Grant XJ2019035, in part by the Anhui Provincial Natural Science Foundation under Grant 1808085J06 and 1908085QF271, and in part by the State Key Laboratory of Synthetical Automation for Process Industries under Grant PAL-N201805 {\it (Corresponding author: Xiaoshu Xiang and Xingyi Zhang).}}
}

\author{Shangshang Yang,
        Ye Tian,
        Xiaoshu Xiang,
        Shichen Peng,
       and
       Xingyi Zhang, \emph{Senior Member, IEEE}
\thanks{S. Yang, S. Peng and X. Zhang is with the Key Laboratory of Intelligent Computing and Signal Processing of Ministry of Education,
School of Artificial Intelligence, Anhui University, Hefei 230039, China (email: yangshang0308@gmail.com; severus.peng@gmail.com; xyzhanghust@gmail.com).}
\thanks{Y. Tian and X. Xiang are with the Key Laboratory of Intelligent Computing and Signal Processing of Ministry of Education, Institutes of Physical Science and Information Technology, Anhui University, Hefei 230601, China (email: field910921@gmail.com; 94xxs13@gmail.com).}
}

\markboth{IEEE Transactions on Neural Networks and Learning Systems}
{Yang \MakeLowercase{\textit{et al.}}: No title}

\IEEEpubid{0000--0000/00\$00.00~\copyright~0000 IEEE}

\maketitle

\begin{abstract}
    Evolutionary neural architecture search (ENAS) has recently received increasing attention by effectively finding high-quality neural architectures,
   which however consumes high computational cost by training the architecture encoded by each individual for complete epochs in individual evaluation.
  Numerous ENAS approaches have been developed to reduce the evaluation cost,
  but it is often difficult for most of these approaches to achieve high evaluation accuracy.
   	To address this issue, in this paper we propose an accelerated ENAS via multi-fidelity evaluation termed MFENAS,
   	where the individual evaluation cost is significantly reduced by training the architecture encoded by each individual for only a small number of epochs.
    The balance between evaluation cost and evaluation accuracy is well maintained by suggesting a multi-fidelity evaluation,
    which identifies the potentially good individuals that cannot survive from previous generations by integrating multiple evaluations under different numbers of training epochs.
    For high diversity of neural architectures, a population initialization strategy is devised to produce different neural architectures varying from ResNet-like architectures to Inception-like ones.
    Experimental results on CIFAR-10 show that the architecture obtained by the proposed MFENAS achieves a 2.39\% test error rate at the cost of only 0.6 GPU days on one NVIDIA 2080TI GPU,
    demonstrating the superiority of the proposed MFENAS over state-of-the-art NAS approaches in terms of both computational cost and architecture quality.
    The architecture obtained by the proposed MFENAS is then transferred to CIFAR-100 and ImageNet,
    which also exhibits competitive performance to the architectures obtained by existing NAS approaches.
    The source code of the proposed MFENAS is available at {\url{ https://github.com/DevilYangS/MFENAS/}}.
\end{abstract}


\section{Introduction}
\IEEEPARstart{D}{eep} learning has made significant success in tackling various machine learning tasks such as classification~\cite{niepert2016learning}, object detection~\cite{liu2016ssd}, speech recognition~\cite{abdel2014convolutional}, natural language processing~\cite{Young2018Recent}, among many others.
For achieving desirable performance, it is widely recognized that the architecture of deep neural networks (DNNs) is very crucial and a number of researchers are devoted to manually designing various neural architectures.
Examples of this category include the ResNet~\cite{he2016deep}, DenseNet~\cite{huang2017densely} and GoogLeNet~\cite{szegedy2015going,szegedy2016rethinking}.
Nevertheless, the hand-craft designed architectures often require extensive human expertise in both DNNs and the problem to be addressed,
which is unrealistic for common users.
This challenge arises a hot research direction in machine learning, called neural architecture search (NAS).

The NAS task was first reported by Zoph and Le in~\cite{zoph2016neural}, where the reinforcement learning was adopted as the optimization algorithm to automatically search for the architecture of convolutional neural networks (CNNs).
Based on the work in~\cite{zoph2016neural}, another famous NAS approach named NASNet was developed by Zoph and Le for CNNs in~\cite{zoph2018learning},
which can obtain neural architectures with extremely competitive performance compared to the hand-craft ones.
Basically, the task of NAS can be considered as the problem of combination optimization of numerous components in DNNs,
which is often computationally very hard to be addressed due to the large search space.
In the past years, a variety of optimization techniques have been adopted for NAS to yield high-performance architectures,
including reinforcement learning (RL)~\cite{pham2018efficient,chen2019renas}, gradient optimization~\cite{liu2018darts,luo2018neural}, Bayesian optimization~\cite{kandasamy2018neural}, and evolutionary algorithm (EA)~\cite{lu2019nsga,sun2020automatically}.

\IEEEpubidadjcol
Among existing optimization techniques for NAS, EAs have attracted increasing attention due to their powerful abilities in dealing with various complex application problems~\cite{prins2004simple,tian2019fuzzy,xiang2020clustering,zhang2019indexed}.
The branch of NAS using EAs is usually called evolutionary neural architecture search (ENAS) and
a number of ENAS approaches have been reported in the past three years~\cite{xie2017genetic,real2017large,liu2018hierarchical,real2019regularized,lu2019nsga,sun2019completely,sun2020automatically}.
For example, Real \emph{et al.}~\cite{real2017large} proposed a large-scale evolution approach to search for the architecture of DNNs by taking about 2750 GPU days, by which the best architecture achieved a 5.4\% error rate on CIFAR-10 dataset~\cite{krizhevsky2009learning};
In~\cite{liu2018hierarchical}, a novel hierarchical genetic representation was devised for EAs to encode the architecture of DNNs,
which can search for the architecture having a 3.75\% error rate within 300 GPU days;
Xie and Yuille~\cite{xie2017genetic} suggested an EA named Genetic-CNN to only optimize the topology of CNNs,
where the best architecture found in 17 GPU days achieved a 7.1\% error rate;
Sun \emph{et al.}~\cite{sun2019completely} suggested an AE-CNN algorithm to evolve CNN architectures based on two types of effective blocks and it took 27 GPU days to obtain an optimized CNN architecture with a 4.3\% error rate.

In spite of the promising performance of found architectures, ENAS usually suffers from high computation cost, making it impractical for real-world applications.
On the one hand, EAs are a population based stochastic algorithm, where numerous individuals in the population need to be evaluated at each generation.
On the other hand, the evaluation of each individual is computationally expensive for the task of NAS since the architecture encoded by the individual requires to be trained for complete epochs.
To address this issue, a variety of approaches have been developed for ENAS,
which can be roughly divided into the following two categories.
The first category of approaches is developed based on the idea of reducing the number of individual evaluations in the process of optimization,
where the surrogate model~\cite{swersky2014freeze,baker2017accelerating,deng2017peephole,sun2019surrogate},
early stopping~\cite{sun2018particle},
population reduction and memory~\cite{assunccao2019fast,sun2020automatically} are three commonly used techniques.
The other category of approaches focuses on reducing the computational cost of training the architecture encoded by each individual in individual evaluation, including weight inheritance~\cite{zhang2020sampled}, subset training~\cite{liu2019deep} and weight sharing~\cite{yang2020cars}.

Nevertheless, it still poses a great challenge for existing accelerated ENAS approaches to achieve high evaluation accuracy when reducing evaluation cost,
which hinders these approaches from finding high-quality architectures at low computational cost.
In this paper, we propose an accelerated ENAS via multi-fidelity evaluation, termed MFENAS, to achieve both low evaluation cost and high evaluation accuracy in NAS.
The proposed MFENAS uses a small number of training epochs in individual evaluation to achieve low evaluation cost,
and employs multi-fidelity evaluation to balance evaluation cost and evaluation accuracy in NAS.
The main contributions of this paper are summarized as follows:

\begin{enumerate}
\item  A novel idea of reducing evaluation cost of individuals is suggested to accelerate ENAS,
where the architecture encoded by an individual is trained for only a small number of epochs rather than complete epochs.
For balancing evaluation cost and evaluation accuracy, we further propose a multi-fidelity evaluation to identify potentially good individuals that cannot survive from previous generations,
where the idea of fidelity is introduced from computational fluid dynamics simulations~\cite{wang2017generic}.
The proposed multi-fidelity evaluation combines low-fidelity evaluation and high-fidelity evaluation that are distinguished by the number of training epochs for individual evaluation.
In this case, the proposed multi-fidelity evaluation takes the advantages of both low computational cost in low-fidelity evaluation and high evaluation accuracy in high-fidelity evaluation, ensuring the balance between evaluation cost and evaluation accuracy.
Hence, the proposed multi-fidelity evaluation can facilitate ENAS in achieving high-quality architectures at low computational cost.
\item
        Based on the proposed multi-fidelity evaluation, a multi-objective EA called MFENAS is developed in the framework of NSGA-II~\cite{deb2002fast} for fast neural architecture search.
        The validation error rate and model complexity are adopted as two optimization objectives with the aim of finding high-accuracy neural architectures under different levels of complexity.
        For high diversity of initial neural architectures, a population initialization strategy is suggested to produce a set of neural architectures varying from Inception-like architectures to ResNet-like ones.
        For maintaining diversity of neural architectures during NAS, a variable-length integer vector is utilized to encode neural architectures, and a genetic operator is devised to equip with variable-length individuals in the proposed MFENAS.

\item
The performance of the proposed MFENAS is verified on three widely used datasets, namely, CIFAR-10, CIFAR-100 and ImageNet.
The search of neural architectures is implemented on CIFAR-10 and the architectures found by the proposed MFENAS are evaluated on CIFAR-10, CIFAR-100 and ImageNet.
    The best architecture found by the proposed MFENAS achieves a 2.39\% test error rate on CIFAR-10 at the cost of only 0.6 GPU days on one NVIDIA 2080TI GPU, which is superior over existing EA and non-EA based NAS approaches.
    The best architecture is transferred to CIFAR-100 and ImageNet, which also holds a promising performance in comparison with existing NAS approaches.
\end{enumerate}

The rest of this paper is organized as follows.
Section~\ref{sec:relatedwork} presents preliminaries and related work on ENAS.
Section~\ref{sec:algorithm} elaborates the proposed MFENAS and
Section~\ref{sec:exp} reports the experimental results for verifying the performance of the proposed MFENAS.
Finally, conclusions and future work are given in Section~\ref{sec:Conclusion}.

\section{Preliminaries and Related Work}\label{sec:relatedwork}

In this section, we first briefly introduce some preliminaries of ENAS.
Then, we review existing work on reducing the computation cost of ENAS.
Finally, we present the motivation of this study.

\subsection{Evolutionary Neural Architecture Search}
Given a data set $D=\{D_{train},D_{val},D_{test}\}$, the neural architecture search can be formulated as
the following single-objective optimization problem.
\begin{equation}
\min_{\mathbf{x}} \mathbf{F}(\mathbf{x}) = 1-f_{val\_accruacy}(\mathbf{x},D),
\end{equation}
where $\mathbf{x}$ represents the architecture to be optimized, $f_{val\_accruacy}(\mathbf{x},D)$
denotes the validation accuracy of the architecture $\mathbf{x}$ on $D_{val}$ after being trained for complete epochs on the training dataset $D_{train}$.

Due to the enormous demand for small-sized networks in real-world applications,
the model complexity is also considered in the NAS task~\cite{elsken2018efficient,lu2019nsga},
and thus NAS can be formulated as a bi-objective optimization problem as follows.
\begin{equation}
\min_{\mathbf{x}} \mathbf{F}(\mathbf{x}) = \left\{
\begin{aligned}
f_1(\mathbf{x})&=1-f_{val\_accruacy}(\mathbf{x},D)\\
f_2(\mathbf{x})&=f_{complexity}(\mathbf{x})\\
\end{aligned}
\right.,\label{compute_fitness}
\end{equation}
where $f_1(\mathbf{x})$ is the classification error rate of $\mathbf{x}$ on validation dataset $D_{val}$
and $f_2(\mathbf{x})$ denotes the model complexity of $\mathbf{x}$.
The model complexity can be measured by the number of model parameters,
the number of floating-point operations and inference time~\cite{lu2019nsga,elsken2018efficient}.
In this paper, we adopt the bi-objective optimization problem formulation for NAS,
where the number of model parameters is used to measure the model complexity.

To address  multi-objective optimization problems, a variety of approaches have been developed
based on different ideas, e.g., Bayesian optimization~\cite{laumanns2002bayesian},
simulated annealing~\cite{bandyopadhyay2008simulated}, Tabu search~\cite{pacheco2006tabu} and EAs~\cite{coello2006evolutionary}.
Among these approaches, EAs are the most widely investigated approaches to multi-objective optimization
due to the fact that they maintain a population approximating the set of optimal solutions.
A large number of EAs have been developed to solve multi-objective optimization problems in different research areas,
such as community detection~\cite{tian2019fuzzy}, shelter location problem~\cite{xiang2020clustering},
and vehicle routing problem~\cite{zhang2019hybrid}.

\begin{figure}[t!]
\begin{center}
\includegraphics[width=1\linewidth]{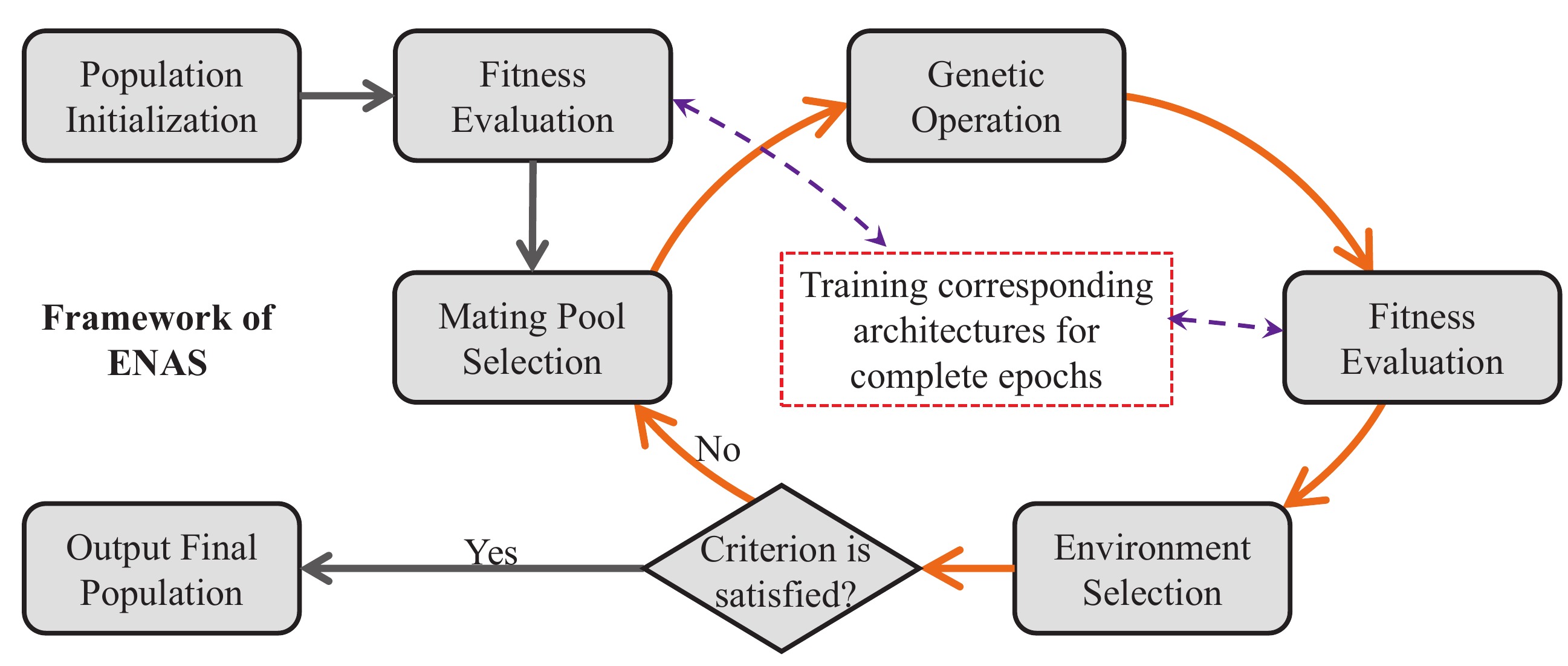}
 \caption{A general framework of EAs for NAS}\label{framework_of_ENAS}
\end{center}
\end{figure}

Generally, an EA consists of five main steps as follows:
(1) Population initialization, by which an initial population is created at the beginning of EA and each individual in the population encodes a solution of the problem to be solved.
(2) Mating pool selection, by which individuals are mated to select parent individuals.
(3) Genetic operation, where crossover and mutation operators are used to create offspring individuals from the parent individuals.
(4) Fitness evaluation, where the quality of each individual in the population is measured.
(5) Environment selection, by which good individuals in the parent and offspring population survive for next generation.
Fig.~\ref{framework_of_ENAS} gives the framework of EAs for NAS tasks.
It is worth noting that the fitness evaluation of each individual for NAS is achieved by training the architecture encoded by each individual for complete epochs on training dataset, which is often characterized with expensive computational cost.

\subsection{Related Work}
To reduce the computational cost of ENAS, many approaches have been developed and
can be roughly divided into the following two categories.

\subsubsection{\textbf{Reducing the Number of Individuals to Be Evaluated in Real}}
Three ideas are often adopted in existing ENAS to reduce the number of individuals to be evaluated in real, namely, surrogate model, early stopping, population reduction and memory.

\textbf{Surrogate Model.}
The main idea of surrogate model based ENAS approaches is to predict the quality of architectures encoded by some individuals by surrogate models to reduce the number of individuals to be evaluated in real.
A representative belonging to this idea is the work reported by Sun \emph{et al.} in~\cite{sun2019surrogate}, where the random forest was adopted as the surrogate model to predict
the quality of newly created individuals at each generation of EAs.
Experimental results indicated that the surrogate model assisted ENAS saved 68.18\% and 69.70\% computational cost on CIFAR10 and CIFAR100, respectively.
Nevertheless, the best architecture found on CIFAR10 can still achieve 94.70\% test accuracy, which is only 0.06\% worse than the version without surrogate model.
There also exist some surrogate assisted NAS approaches which are not based on the framework of EAs~\cite{swersky2014freeze,deng2017peephole}.

\textbf{Early Stopping.}
The idea of early stopping is to terminate the training of poor individuals based on the performance of architectures in the early training stage to reduce the number of individuals to be evaluated.
In~\cite{Suganuma2020Evolution}, Suganuma \emph{et al.} developed an early stopping strategy based on a reference curve to accelerate ENAS,
where the reference curve is updated on the basis of the best individual in population at each generation.
A similar idea is also adopted for accelerating ENAS in~\cite{assunccao2019automatic} and~\cite{so2019evolved}.
Despite that the idea of early stopping can considerably reduce the computational cost of ENAS approaches, it easily leads to inaccurate estimation on individual quality, especially for complicated architectures as stated in~\cite{liu2020survey}.
This makes it hard for early stopping based ENAS approaches to achieve competitive performance in NAS.

\textbf{Population Reduction and Memory.}
The population reduction achieves the accelerating of ENAS by recursively reducing the size of population~\cite{assunccao2019fast,fan2020evolutionary},
whereas the population memory avoids evaluating the same individuals that have been evaluated in the evolution of population to solve this task~\cite{sun2020automatically,johner2019efficient}.
Fan \emph{et al.}~\cite{fan2020evolutionary} divided the whole optimization process of the ENAS approach
into different stages with different population sizes,
where a large population is used to ensure the global search ability in early stage
and a small population is adopted to reduce the number of individuals to be evaluated in population in later stage.
In~\cite{sun2020automatically}, Sun \emph{et al.} utilized a hashing method to
record both architecture information and fitness of each individual,
so that the fitness value for an individual can be directly obtained if this individual has been recorded.
Despite avoiding redundant individual evaluations in ENAS,
these strategies still consume expensive computational cost to achieve high-quality architectures~\cite{liu2020survey}.

\subsubsection{\textbf{Reducing the Evaluation Cost of Each Individual}}\indent
The subset training, weight inheritance and weight sharing are three ideas which are widely adopted in ENAS approaches to reduce the evaluation cost of each individual.

\textbf{Subset Training.}
The subset training is to train the architecture on a small subset selected from the original dataset having a large number of data,
by which the computation cost on evaluating each individual can be effectively reduced.
Following this idea, Liu \emph{et al.}~\cite{liu2019deep} suggested an ENAS approach to search for the optimal architecture for medical image denoising.
To reduce the training time without seriously degrading the performance,
a small subset having similar properties to those in the original dataset is randomly selected for exploring promising CNN architectures.
Despite that the subset training can effectively reduce computational cost of ENAS,
there exists a big challenge of overfitting problem due to the difficulty of properly selecting a subset.

\textbf{Weight Inheritance.}
The weight inheritance is a technique that the architectures encoded by offspring individuals inherit most weights in the architectures encoded by parent individuals,
since the genetic operators in ENAS produce offspring individuals by inheriting most architectures encoded by parent individuals
~\cite{zhang2018finding,zhang2020sampled}.
In this way, the training of architecture encoded by each individual can be accelerated by means of initiating the training with the inherited weights instead of starting from scratch, which accelerates the evaluation cost of each individual.
Zhang \emph{et al.}~\cite{zhang2018finding} adopted the weight inheritance in the evaluation of offspring individuals in ENAS,
where the architectures encoded by these individuals are trained for only one epoch
based on the weights inherited from parent individuals.
In~\cite{zhang2020sampled}, a sampled training strategy was designed to train all parent individuals at the beginning of each iteration,
and the offspring individuals at each iteration can be directly evaluated on a validation dataset without any training based on the weights inherited from the parent individuals.

\textbf{Weight Sharing.}
The weight sharing is to reduce the evaluation cost of each individual by directly obtaining the weights of the architecture encoded by each individual from a set of weights stored in a SuperNet model.
Hence, the individuals in ENAS can be evaluated by training only one SuperNet model instead of training all the architectures encoded by these individuals, which thus considerably reduces the computational cost of ENAS~\cite{pham2018efficient}.
In~\cite{yang2020cars}, Yang \emph{et al.} developed a continuous EA for efficient NAS approach based on weight sharing,
where all individuals generated in NAS share the same set of weights in one SuperNet.
The weights in the SuperNet are updated by training the architectures encoded by non-dominated individuals at each generation of ENAS.

\subsection{Motivation}\label{sec:motivation}
\begin{figure}[t!]
\begin{center}
 \centering
    \subfloat[Validation accuracy of 10 randomly selected architectures.]{
        \includegraphics[width=0.47\linewidth]{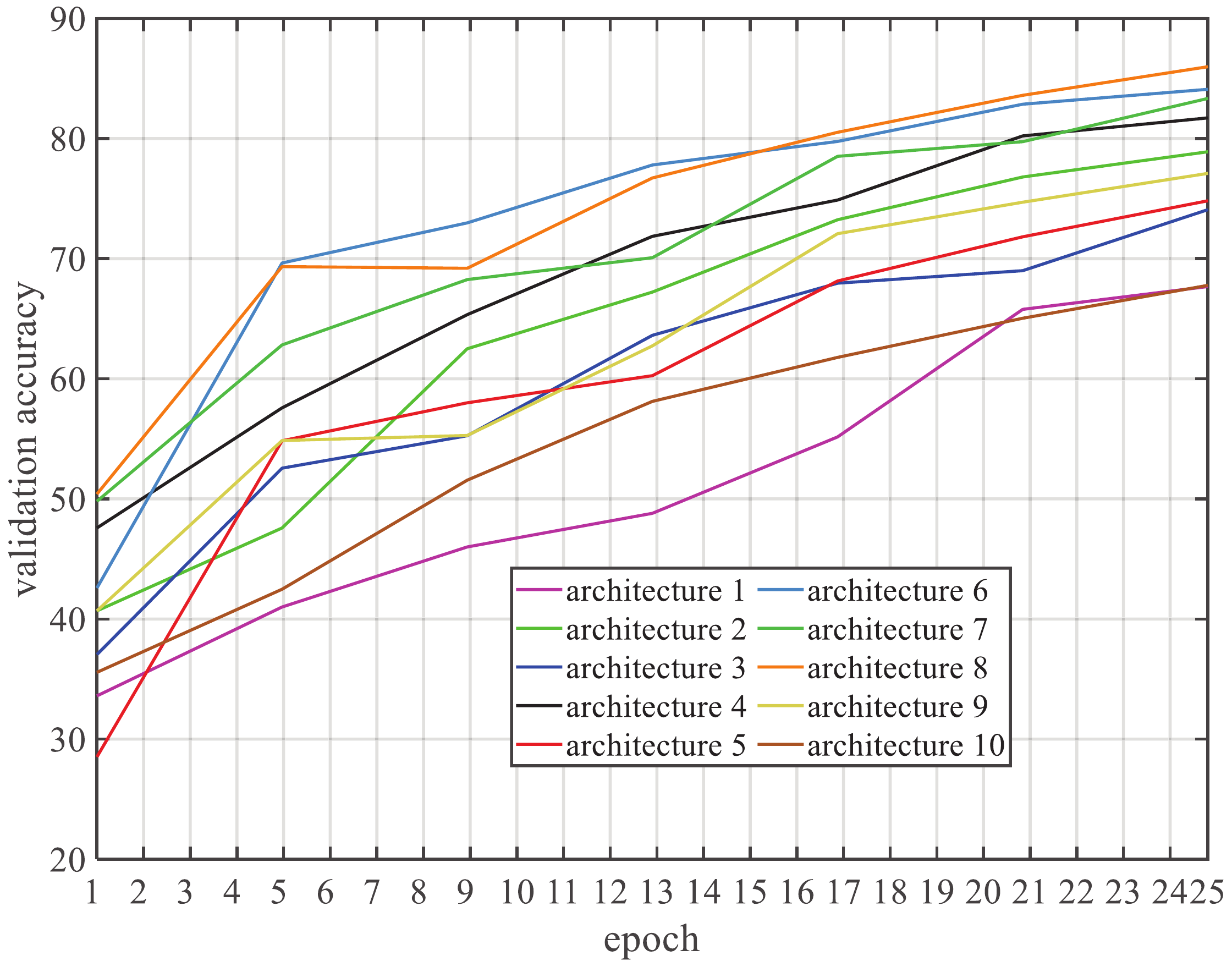}
    }\hfil
    \subfloat[The Kendall Tau Rank Correlation Coefficient on 1000 architectures]{
        \includegraphics[width=0.47\linewidth]{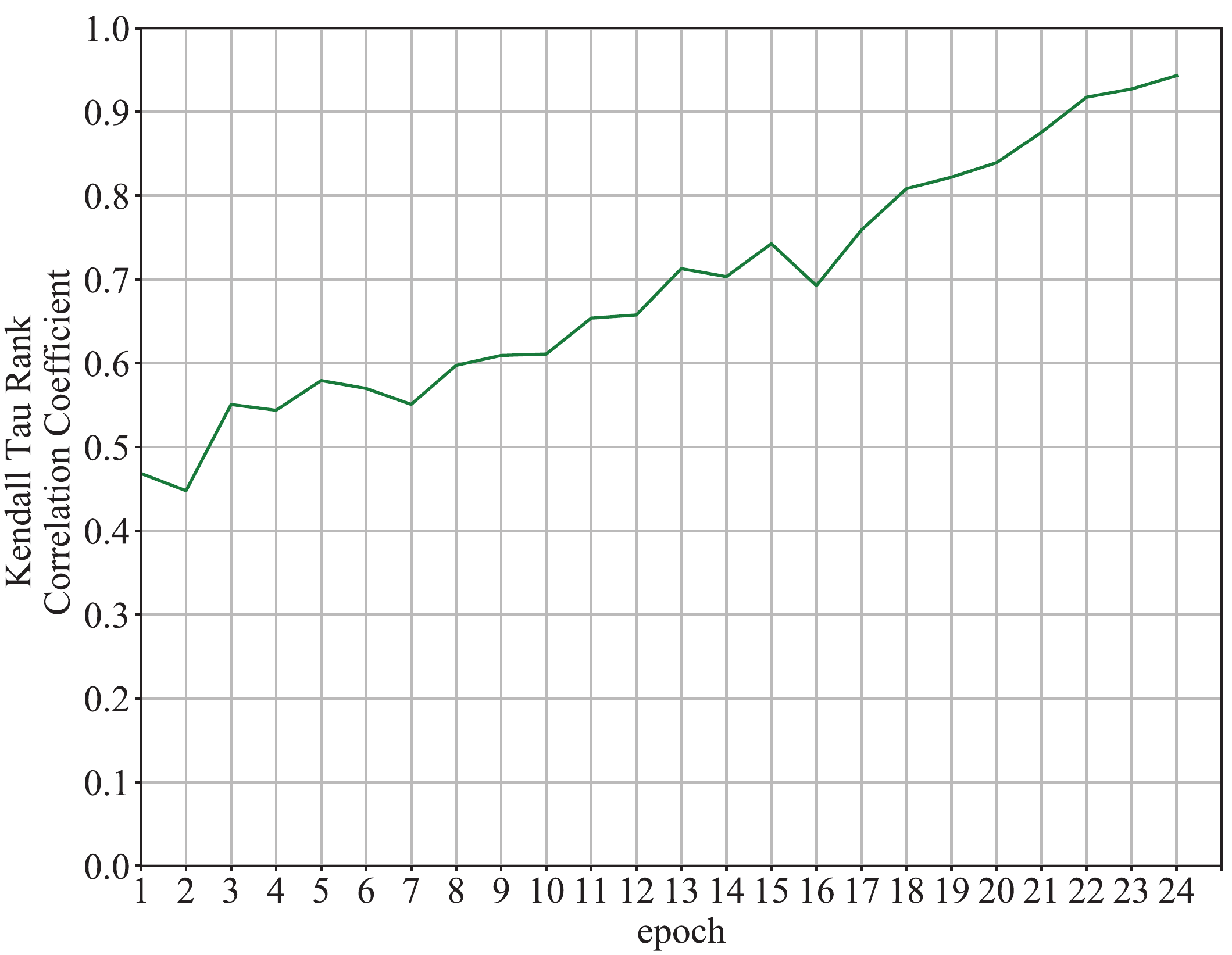}
    }
\end{center}

\caption{The performance ranking observation on 1000 randomly sampled architectures in cell based architecture search space. It is worth noting that the number of complete epochs for training models is set to 25 according to NASNet\cite{zoph2018learning} and NSGA-Net\cite{lu2019nsga}.}\label{accuracy_lines}

\end{figure}

Different from existing accelerated ENAS approaches, the proposed MFENAS reduces the evaluation cost of each individual by means of
training the architecture encoded by each individual for only a small number of epochs rather than complete epochs.
This idea is motivated from the observation that the performance ranking of individuals at each generation will not change considerably no matter whether the architectures encoded by the individuals are trained for a small number of epochs or complete epochs.
To illustrate this fact, Fig.~\ref{accuracy_lines} shows the profiles of performance ranking on 1000 architectures under different numbers of training epochs,
where 1,000 neural architectures are randomly sampled in the cell based search space according to the setting of training suggested in~\cite{zoph2018learning,luo2018neural,lu2019nsga},
and the number of nodes in each cell ranges from $5$ to $12$.
Specifically, Fig.~\ref{accuracy_lines} (a) presents the validation accuracy of 10 architectures that are randomly selected from the 1000 architectures,
where the architectures having high accuracy under a small number of training epochs also hold high accuracy under complete epochs and vice verse.

This observation also often holds when the accuracy difference between two architectures is not significant as depicted in Fig.~\ref{accuracy_lines} (b),
which gives the profile of the Kendall Tau Rank Correlation Coefficient (Kendall's $\tau$ for short) between the performance ranking at each epoch and that of final epoch on all of the 1,000 randomly sampled architectures.
The Kendall's $\tau$ is a widely used indicator to measure the correlation between two different
rankings of items~\cite{Sen1968Estimates}.
The $\tau \in [-1,1]$, where $\tau =1$ means the two rankings are exactly the same and $\tau =-1$ means they are completely opposite.

From Fig.~\ref{accuracy_lines} (b), we can find the following two results.
On the one hand, the performance ranking of architectures at any number of training epochs holds a high correlation with the ranking at final epoch,
where the Kendall's $\tau$ value is larger than 0.45 and competitive in comparison with most of the surrogate models used in ENAS~\cite{swersky2014freeze,deng2017peephole,sun2019surrogate}.
On the other hand, the Kendall's $\tau$ value increases considerably as the number of training epochs increases,
which inspires us to consider the idea of multi-fidelity optimization (MFO)~\cite{wang2017generic} to ensure the performance of ENAS under a small number of epochs.
It is worth noting that the MFO is a popular idea for striking the balance between effectiveness and efficiency of an algorithm
by means of combining high fidelity and low fidelity,
where the high fidelity leads to high-accuracy evaluation but low computation efficiency,
and the low fidelity leads to high computation efficiency but low-accuracy evaluation~\cite{ren2016application}.

\begin{figure*}[t]
\begin{center}
\includegraphics[width=0.85\linewidth]{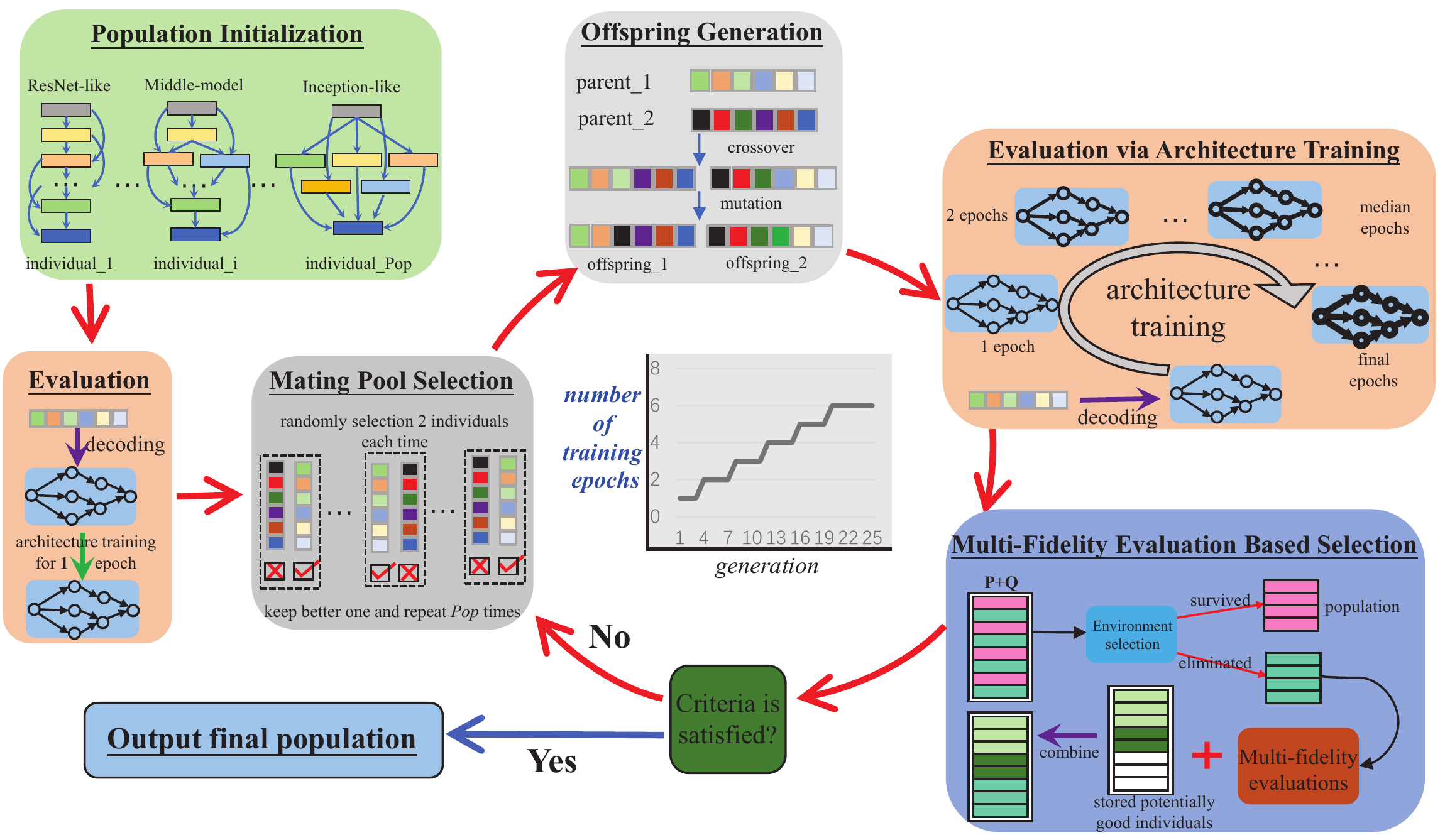}
 \caption{The overall framework of the proposed MFENAS.}\label{fig:new_framework}
\end{center}
\end{figure*}

Following the above idea, we propose an accelerated ENAS approach via multi-fidelity evaluation, named MFENAS,
where each individual is evaluated by training the architecture encoded by each individual for only a small number of epochs.
For balancing evaluation cost and evaluation accuracy, we suggest a multi-fidelity evaluation combing high-fidelity evaluation and low-fidelity evaluation, where high-fidelity evaluation indicates individual evaluation under a large number of training epochs,
and low-fidelity evaluation indicates that under a small number of training epochs.
Hence, the multi-fidelity evaluation can identify potentially promising individuals that cannot survive from previous generations
by integrating multiple evaluations under different numbers of training epochs.

\section{The Proposed MFENAS} \label{sec:algorithm}
In this section, we first present the overall framework of the proposed MFENAS.
Then, we elaborate the multi-fidelity evaluation in the proposed MFENAS.
Finally, further details in the proposed MFENAS are presented.

\subsection{Overall Framework of MFENAS}\label{section_overall_precedure}


To start with, the formal notations used in this paper are listed in Table~\ref{tab:notation}.

To simultaneously optimize the accuracy and complexity of neural architecture in NAS,
the proposed MFENAS employs NSGA-II as the optimizer due to its high competitiveness in solving bi-objective optimization problems~\cite{Lamont2007Evolutionary}.
Compared with existing work on NSGA-II for NAS, the proposed MFENAS is characterized with two key aspects:
a small number of architecture training epochs for individual evaluation,
and a multi-fidelity evaluation for environment selection at each generation.
Specifically, Fig.~\ref{fig:new_framework} presents the overall framework of MFENAS, which mainly consists of six steps.

First, a well-distributed population is initially generated by producing diverse neural architectures that vary from ResNet-like architectures to Inception-like ones.
Second, the individuals in the generated population are efficiently evaluated by training the neural architectures encoded by these individuals for a certain number of epochs, which is initially set to one.
Third, binary tournament selection is applied to the generated population to mate individuals in the population.
Fourth, an offspring population is produced from the parent population that is generated at last step by implementing revised
single-point crossover and mutation (described in Section~\ref{sec:genetic}) in mated individuals of the parent population.
Fifth, the individuals in the offspring population are efficiently evaluated by training the neural architectures encoded by these individuals for a certain number of epochs, which will be updated according to the generation and the number of training epochs.
Finally, a multi-fidelity evaluation based selection is used to select individuals from the union of parent population and offspring population,
where the multi-fidelity evaluation identifies and maintains potentially good individuals that are eliminated by the environment selection in NSGA-II~\cite{deb2002fast}.
The second to the sixth step will repeat until the evolution termination criteria is satisfied,
after which the non-dominated individuals will be output.
For more details, Algorithm~\ref{algorithm: MFENAS} presents main steps of the proposed MFENAS.

\begin{table}[!t]
\renewcommand{\arraystretch}{0.6}
\centering
\footnotesize
\caption{Formal Notations Used in This Paper.}
\label{tab:notation}
\setlength{\tabcolsep}{0.4mm}{

\begin{tabular}{c|l}

\hline
\hline
\makebox[1.1cm][c]{\textbf{Notation}}&\makebox[7cm][c]{\textbf{Description}}\\
\hline
$\mathbf{P}$ ($\mathbf{Q}$)&\ a population $\mathbf{P}=\{\mathbf{P_1},\mathbf{P_2},\cdots,\}$ (a offspring population)\\
\hline
$\mathbf{P}'$ &\ a parent population for generating offsprings\\
\hline
$\mathbf{P^S}$ ($\mathbf{P^E}$)&\ a population surviving (eliminated) from selection\\

\hline
$\mathbf{E}$ &\ an archive storing potentially good individuals\\

\hline
$I$ ($\mathbf{P}_i$) &\ the individual (the $i$-th individual in $\mathbf{P}$) $I = \mathbf{P_i}=\{\mathbf{NC}, \mathbf{RC}\}$\\

\hline
$\mathbf{NC}$ ($\mathbf{RC}$) &\ a normal cell (a reduction cell) $\mathbf{NC} (\mathbf{RC}) = (\mathbf{n_0},\mathbf{n_1},\cdots)$  \\

\hline
$\mathbf{n_i}$ &\ the $i$-th node in $\mathbf{NC} (\mathbf{RC})$, $\mathbf{n_i}=(\mathbf{L}, Op)$\\
\hline
$\mathbf{L}$ & a vector denotes the link information\\
\hline
 $Op$ &\ a number refers to used operation\\

\hline
$Gen$, $Pop$ &\ maximum number of generations, Population size\\

\hline
 $Node_{range}$ &\ range of the number of nodes in initialization\\
 \hline
 $MF$&\  number of fidelities for multi-fidelity evaluation\\
 \hline
 $S$&\  current number of epoch for training process\\
 \hline
 $SS_{counter}$ &\  count of surviving according to one evaluation\\
 \hline
 $MS_{counter}$ &\   count of surviving according to multi-fidelity evaluation\\
 \hline
$ME_{counter}$ &\  count of being eliminated according to multi-fidelity evaluation\\

\hline
\hline

 \multirow{2}[1]{*}{$H$}&\ all previous outputs before node $k$\\
 & \ $H=\{\widehat{h}[0], \widehat{h}[1], h[0], ..., h[k-1] \}$\\
\hline

$\hat{h}[0]$, $\hat{h}[1]$ &\  the outputs of two previous cells\\

\hline
$h[k]$, $in_i$ &\ the output of node $k$, the $i$-the input of a node\\

\hline
\hline

\end{tabular}

}
\end{table}

\begin{algorithm}[t!]
\caption{Main Steps of MFENAS}
\label{algorithm: MFENAS}
\footnotesize
\begin{algorithmic}[1]

\REQUIRE
{$Gen$: Maximum number of generations;
 $Pop$: Population size;
 $Node_{range}$: Range of the number of nodes in initialization;
 $MF$:  Number of fidelities for multi-fidelity evaluation;
   }
\ENSURE
$\mathbf{P}$: Population;
\STATE $\mathbf{P},\mathbf{E},S \leftarrow Initialization(Pop,Node_{range}),\emptyset,1$;   \% Algorithm~\ref{algorithm_Initialization}

\STATE  Evaluate individuals in $\mathbf{P}$ by training the architectures encoded by the individuals for $S$ epoch and compute the fitness by (\ref{compute_fitness});

\FOR{$g=1$ to $Gen$}
\STATE $\mathbf{P}' \leftarrow$ Select $Pop$ parents according to the non-dominated front number and crowding distance of individuals in $\mathbf{P}$; \\ \qquad \qquad \qquad \qquad \qquad \qquad \qquad \quad  \% Mating pool selection

\STATE $\mathbf{Q} \leftarrow Genetic\ Operator(\mathbf{P}')$; \qquad   \qquad   \quad   

\STATE  Evaluate individuals in $\mathbf{Q}$ by training the architectures encoded by the individuals for $S$ epoch(s) and compute the fitness by (\ref{compute_fitness});

\STATE $\mathbf{P},\mathbf{E},S \leftarrow$ $Multi$-$Fidelity\ Evaluation\ Based \ Selection$
        \\\qquad \qquad $(\mathbf{P},\mathbf{Q}, \mathbf{E},g,MF,S)$;  \qquad \qquad \quad    \% Algorithm~\ref{algorithm_MFB}

\ENDFOR
\RETURN $\mathbf{P}$;
\end{algorithmic}
\end{algorithm}

\subsection{The Proposed Multi-Fidelity Evaluation}
In the proposed MFENAS, it is expected that the evaluation cost of individuals can be significantly reduced by training the architecture encoded by each individual for only a small number of epochs,
which will greatly accelerate the entire ENAS process.
Nevertheless, the evaluation cost reduction hinders the architecture training from achieving high accuracy in the individual evaluation,
and thus leads to inappropriate elimination of good individuals in the environment selection~\cite{sun2019surrogate,zhang2020sampled,liu2019deep}.
To address this issue,
we propose a multi-fidelity evaluation based selection for maintaining potentially good individuals that cannot survive from the current environment selection
by multiple evaluations under different numbers of training epochs for each of these individuals.

\begin{figure}[t!]
\begin{center}
\centering
        \includegraphics[width=1\linewidth]{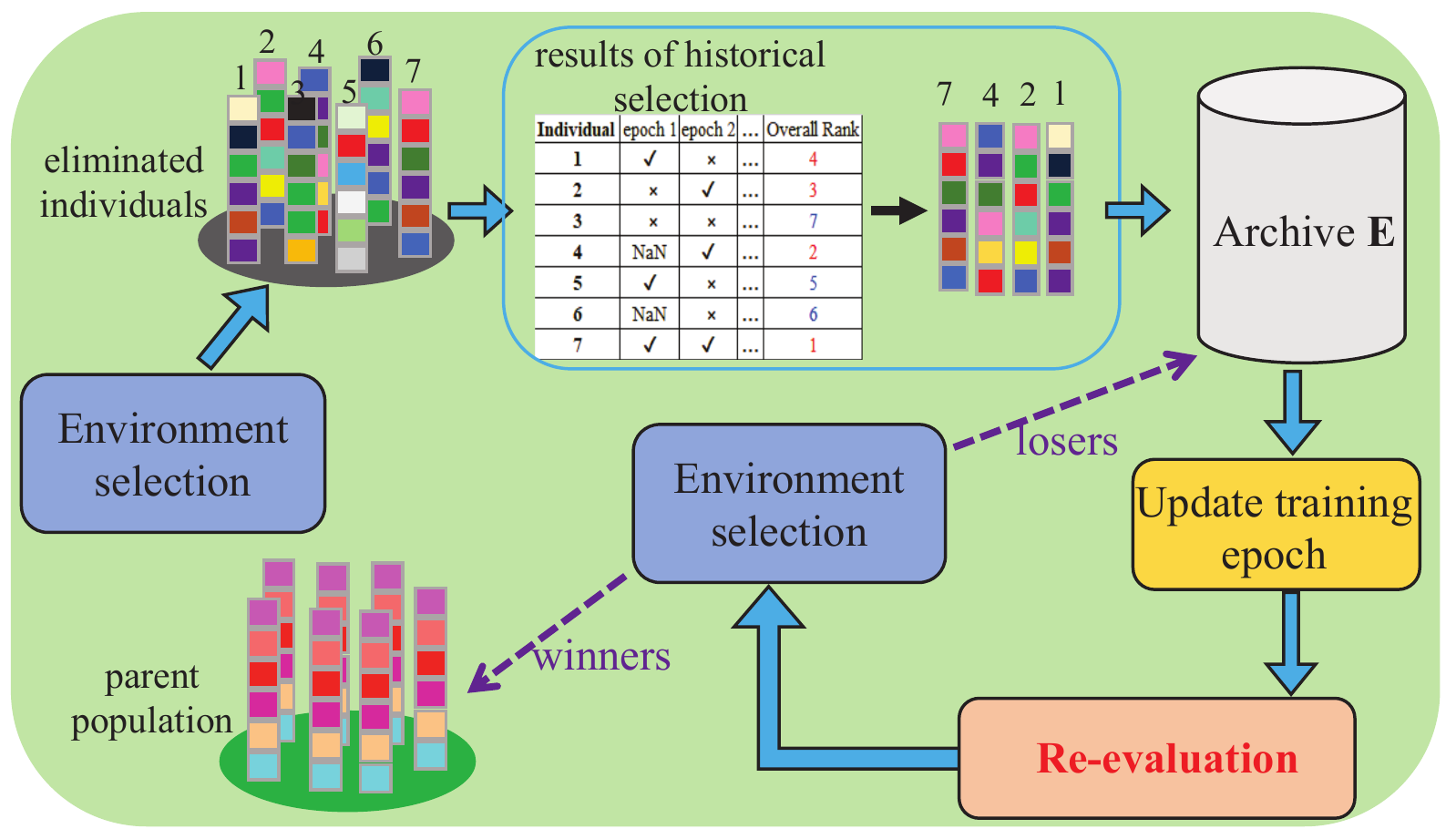}
 \caption{The main procedure of multi-fidelity evaluation based selection.}\label{fig:MF}
\end{center}
\end{figure}

The proposed multi-fidelity evaluation based selection mainly consists of five steps as shown in Fig.~\ref{fig:MF}.
At each generation of ENAS, the environment selection of NSGA-II is first implemented to select a set of individuals from parent and offspring populations as mentioned in Section~\ref{section_overall_precedure}, and to eliminate the remaining individuals in the two populations.
Second, the potentially good individuals are selected from the eliminated individuals by multi-fidelity evaluation,
which are stored in an archive that includes previous potentially good individuals and their survival or elimination information in previous environment selection.
Third, the number of training epochs updates according to the generation and the number of training epochs.
Fourth, the updated number of training epochs is adopted in training architectures encoded by the individuals in the archive and those surviving in the environment selection,
and thus these individuals continue to be re-evaluated based on their model parameters obtained from previous training.
Finally, the environment selection in NSGA-II is applied to the re-evaluated individuals to maintain some good individuals as the parent population at next generation,
while the remaining individuals in the selection will be stored in the archive.

Specifically, the detailed steps of the proposed multi-fidelity evaluation based selection are presented in Algorithm~\ref{algorithm_MFB}.
To begin with, the environment selection of NSGA-II is applied to the union of parent population $\mathbf{P}$ and offspring population $\mathbf{Q}$,
where surviving individuals and eliminated individuals in the selection are denoted as $\mathbf{P^{S}}$ and $\mathbf{P^{E}}$, respectively (Step~\ref{mfb:eselect0}).

Then a multi-fidelity evaluation is employed to select potentially good individuals from $\mathbf{P^{E}}$
by making use of individual selection under different numbers of epochs for training the architectures encoded by the individuals.
To record the individual selection under different numbers of training epochs,
we devise three counters for each individual in $\mathbf{P}$ and $\mathbf{Q}$:
\begin{itemize}
\item The $SS_{counter}$ denotes the count of surviving according to one evaluation;
\item The $MS_{counter}$ denotes the count of surviving according to multi-fidelity evaluation;
\item The $ME_{counter}$ denotes the count of being eliminated according to multi-fidelity evaluation.
\end{itemize}

Based on the environment selection, the $SS_{counter}$ adds one for each individual in $\mathbf{P^{S}}$,
and the $ME_{counter}$ adds one for each individual in $\mathbf{P^{E}}$ whose $ME_{counter}$ is zero (Step~\ref{mfb:counter0}).

\begin{algorithm}[t!]
\caption{Multi-Fidelity Evaluation Based Selection
$(\mathbf{P},\mathbf{E}, \mathbf{E}',g,MF,S)$}
\label{algorithm_MFB}
\footnotesize
\begin{algorithmic}[1]

\REQUIRE
{$\mathbf{P}$: Population;
$\mathbf{Q}$: Offspring;
$\mathbf{E}$: Archive;
$g$: Current generation;
$MF$: Number of fidelities for multi-fidelity evaluation;
$S$: Current number of training epochs;
   }
\ENSURE
$\mathbf{P}$: Population;
$\mathbf{E}$: Archive;
\STATE $\mathbf{P^{S}}, \mathbf{P^{E}} \leftarrow $ Select $Pop$ individuals from $\mathbf{P}+ \mathbf{Q}$ by the environment selection of NSGA-II;\label{mfb:eselect0}

\STATE $SS_{counter}$ +1 for individuals in $\mathbf{P^{S}}$,\\ $ME_{counter}$ +1 for individuals in $\mathbf{P^{E}}$ whose $ME_{counter}$ == 0;\label{mfb:counter0}
\STATE $\mathbf{T} \leftarrow$ Sort the individuals in $\mathbf{E}$ and $\mathbf{P^{E}}$ using four criteria;\label{mfb:prisort}
\STATE $\mathbf{E} \leftarrow$ \emph{Truncated operation} ($\mathbf{T}$);\label{mfb:trunc}

\IF{$mod(g, \lfloor \frac{Gen}{MF}\rfloor)==0$ and $S\neq MF$}
    \STATE $S \leftarrow S+1$;
    \STATE Continue training architectures encoded by individuals in $\mathbf{P^{S}}+ \mathbf{E}$ for $S$ epochs to
        re-evaluate these individuals by (\ref{compute_fitness});\label{mfb:reevaluate}
    \STATE $\mathbf{P},\mathbf{E} \leftarrow $ Select $Pop$ individuals from $\mathbf{P^{S}}+ \mathbf{E}$ by the environment selection of NSGA-II;\label{mfb:eselect1}
    \STATE $MS_{counter}$ +1, $SS_{counter}$ = 0 for individuals in $\mathbf{P}$, \\
     $MS_{counter}$ +1 for individuals in $\mathbf{E}$ whose $ME_{counter}$ == 0,\\
     $ME_{counter}$ +1 for individuals in $\mathbf{E}$;\label{mfb:counterup1}
     \ELSE
     \STATE $\mathbf{P} \leftarrow \mathbf{P^S}$;
\ENDIF

\IF{$g==Gen$}
    \STATE Continue training architectures encoded by individuals in $\mathbf{P^{S}}+\mathbf{E}$ for complete epochs to get final
    validation error and number of model parameters in (\ref{compute_fitness});\label{mfb:reevaluate2}
    \STATE $\mathbf{P},\mathbf{E} \leftarrow $ Select $Pop$ individuals from $\mathbf{P^{S}}+ \mathbf{E}$ by the environment selection of NSGA-II;\label{mfb:eselect2}
\ENDIF

\RETURN $\mathbf{P}$, $\mathbf{E}$, $S$;
\end{algorithmic}
\end{algorithm}

Based on the three counters, four criteria is suggested to sort
the individuals in the union of $\mathbf{E}$ and $\mathbf{P^{E}}$ (Step~\ref{mfb:prisort}):
\begin{equation}\label{equation_rules}
\small
\left\{
\begin{aligned}
&\ MS_{counter}-ME_{counter}\\
&\ -(MS_{counter}+ME_{counter})\\
&\ SS_{counter}\\
&\ f_2(\mathbf{x}) \nonumber\\
\end{aligned}
\right.,
\end{equation}
where the first criteria holds the highest priority and aims to retain the individuals that frequently survive according to multi-fidelity evaluation,
the second criteria holds the second highest priority and tends to maintain the individuals that rarely experience evaluations under different fidelities,
the third criteria holds the third highest priority and is prone to keep the individuals that frequently survive according to one evaluation,
and the fourth criteria holds the fourth highest priority and prefers the individuals that are encoded from large and complex neural architectures.
The individuals in the union of $\mathbf{E}$ and $\mathbf{P^{E}}$ are sorted according to the four criteria.
Next, a truncated operation is used to select the former $\mathbf{|E|}$ highest-criterion individuals and store these individuals to the archive $\mathbf{E}$,
where $\mathbf{|E|}$ denotes a predefined size of $\mathbf{E}$ (Step~\ref{mfb:trunc}).

To further evaluate the individuals in archive $\mathbf{E}$, we periodically update the number of training epoch $S$ for the architectures encoded by the individuals by adding 1 to $S$ for every $\lfloor \frac{Gen}{MF}\rfloor$ generations,
where $Gen$ is the maximum number of generations and $MF$ is the number of fidelities for multi-fidelity evaluation
(also denoting the maximum number of training epochs for each individual evaluation).
With the updated number of training epochs $S$,
the individuals in both archive $\mathbf{E}$ and population $\mathbf{P^{S}}$ are re-evaluated
by continuing training the architectures encoded by these individuals for $S$ epochs based on model parameters obtained from previous training (Step~\ref{mfb:reevaluate}).

After the re-evaluation, the environment selection of NSGA-II is applied to the union of $\mathbf{P^{S}}$ and $\mathbf{E}$ to obtain
a surviving population for a new parent population and an eliminated population stored in the archive $\mathbf{E}$ at next generation (Step~\ref{mfb:eselect1}).
The three counters are then updated according to environment selection:
the $MS_{counter}$ adds one, the $SS_{counter}$ is set as zero for each individual in $\mathbf{P}$,
the $MS_{counter}$ adds one for the individuals that are stored in archive $\mathbf{E}$ for the first time,
and the $ME_{counter}$ adds 1 for each individual in archive $\mathbf{E}$ (Step~\ref{mfb:counterup1}).

When it achieves the maximum number of generations,
all architectures encoded by the individuals in the union of $\mathbf{P^{S}}$ and $\mathbf{E}$ continue to be trained for complete epochs to
get their final fitness values in (\ref{compute_fitness}) (Step~\ref{mfb:reevaluate2}).
Based on the fitness values, the environment selection of NSGA-II is used to
acquire a final population consisting of non-dominated individuals (Step~\ref{mfb:eselect2}).

Note that the number of fidelities for multi-fidelity evaluation $MF$ refers to
how many different types of evaluations in the proposed MFENAS,
and one type of evaluation indicates an individual evaluation under a specific number of training epochs.
For examples, $MF=1$ indicates that only one type of evaluation is used for each individual,
i.e., architecture training for one epoch.
When $MF=3$, it indicates that three types of evaluations can be used for one individual,
i.e., architecture training for one epoch, two epochs and three epochs.
A large value of $MF$ leads to an accurate individual evaluation,
while a small value of $MF$ leads to an efficient individual evaluation for the proposed MFENAS.
Hence, $MF$ plays an important role in striking a balance between evaluation efficiency and accuracy,
which will be discussed in the experiments of Section~\ref{sec:exp_validation}.

\subsection{Related Details}
In this section, we give further details of the proposed MFENAS, including architecture search space,
encoding strategy, genetic operator and population initialization strategy.

\subsubsection{\textbf{Architecture Search Space for NAS}}\label{sec:sspace}\indent

\begin{figure}[t!]
\begin{center}
\includegraphics[width=0.9\linewidth]{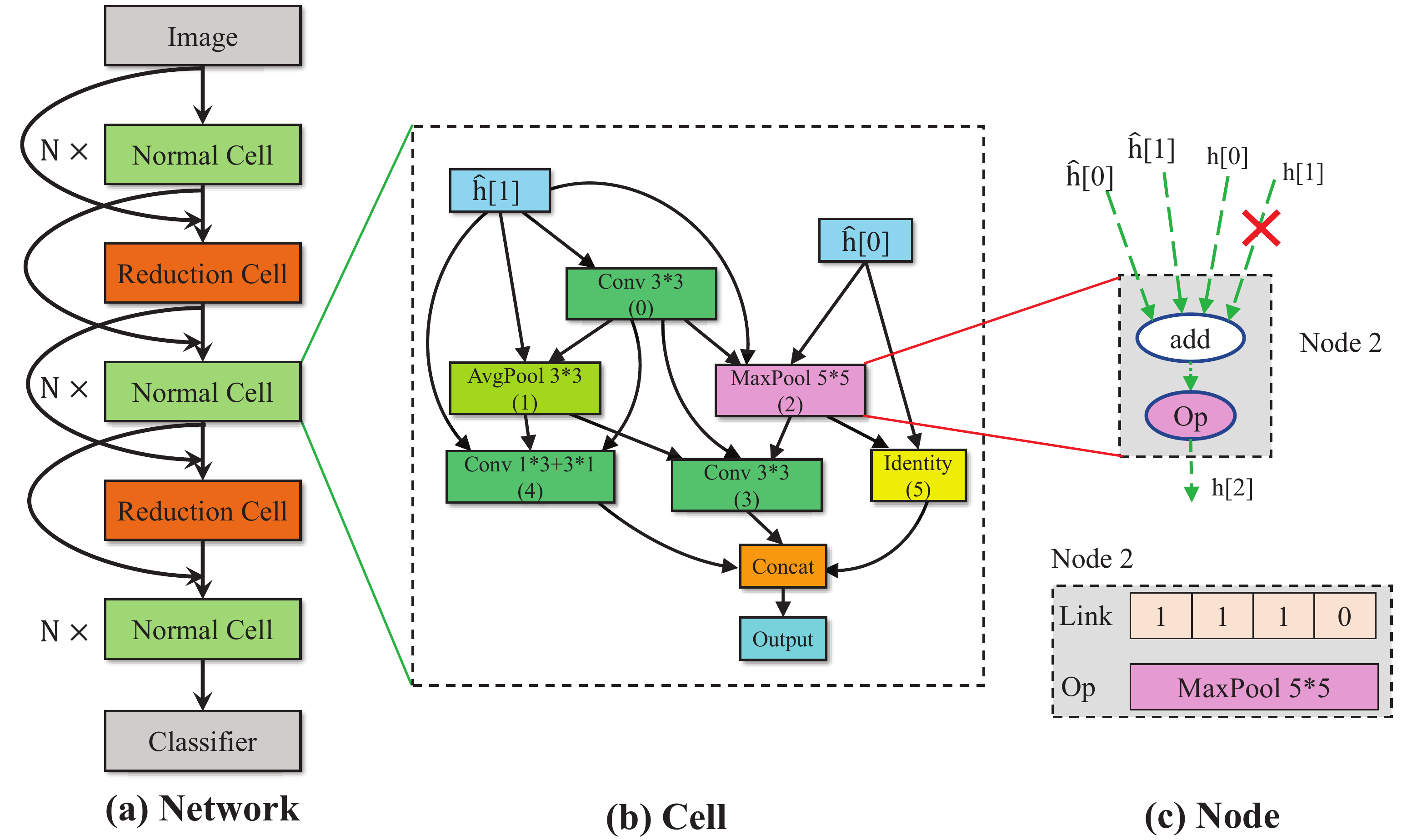}
 \caption{The architecture search space used for the proposed MFENAS.}\label{search_space}
\end{center}
\end{figure}

\textbf{Network.}
As illustrated in Fig.~\ref{search_space} (a), the cell based architecture search space~\cite{zoph2018learning} is adopted to build the whole network,
which consists of a stack of several cells including normal cells and reduction cells.
Either a normal cell or a reduction cell regards the outputs of two previous cells as input.
All the normal cells share the same architecture but different weights.
Similarly, all the reduction cells share the same architecture but different weights.

\textbf{Cell.}
Each cell can be defined as a convolutional network mapping two $H\times W\times F$ tensors (transformed from the two previous cells) to one $H'\times W'\times F'$ tensor.
According to the study in~\cite{elsken2019neural},
operations with stride 1 are used for the normal cell, where $H'=H$, $W'=W$ and $F'=F$.
Reduction cell applies these operations with stride 2,  where $H'=H/2$, $W'=W/2$ and $F'=2F$.
As illustrated in Fig.~\ref{search_space} (b), a normal cell or a reduction cell can be considered as a directed acyclic graph consisting of several nodes,
including two input nodes, one output node, $k$ ($k=6$ in Fig.~\ref{search_space} (b)) computation nodes and one concatenate node (denoted as Concat).
For these nodes, the outputs that are not used will be concatenated together for final output of one cell.

\begin{figure}[t!]
\begin{center}
    \centering
    \subfloat[Node (block) in NASNet]{
        \includegraphics[width=0.2\linewidth]{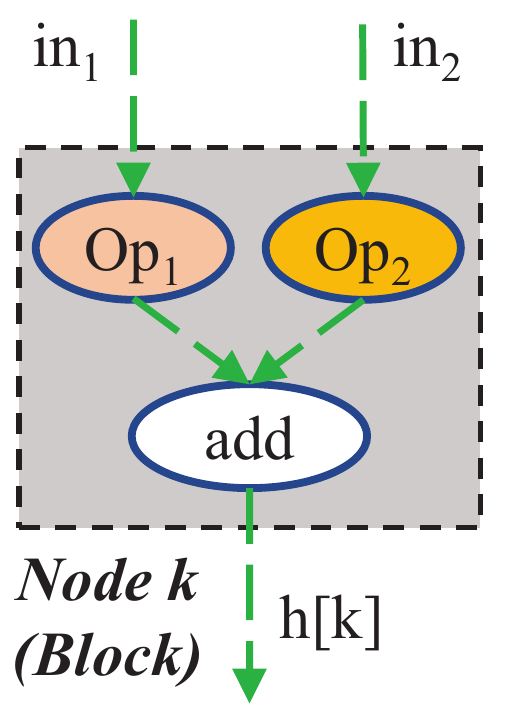}
    }\hfil
    \subfloat[Node in one-shot Approaches]{
        \includegraphics[width=0.4\linewidth]{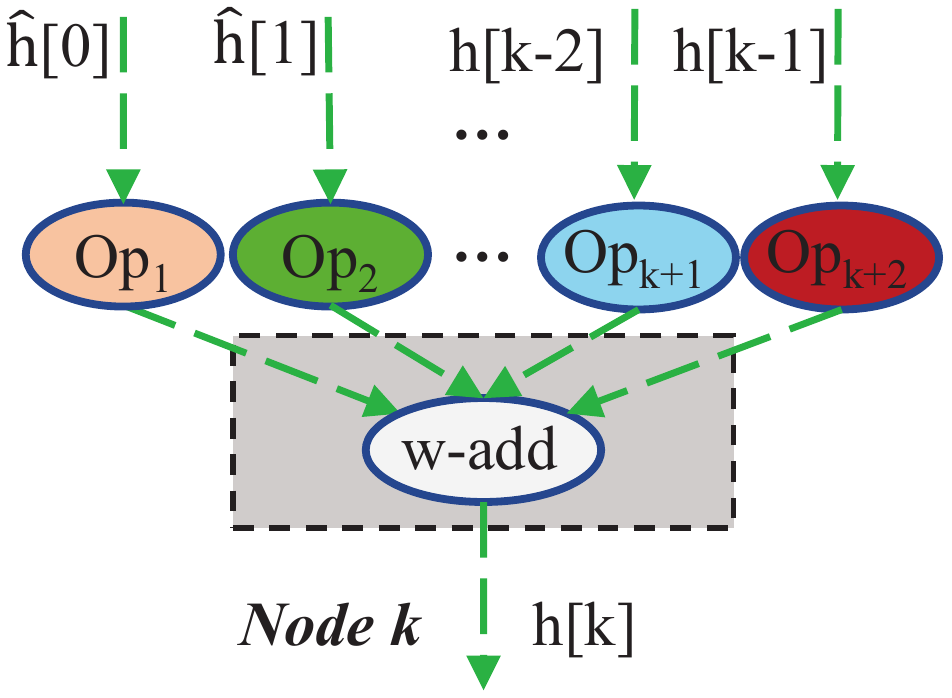}
    }\hfil
    \subfloat[Node in our approach ]{
        \includegraphics[width=0.25\linewidth]{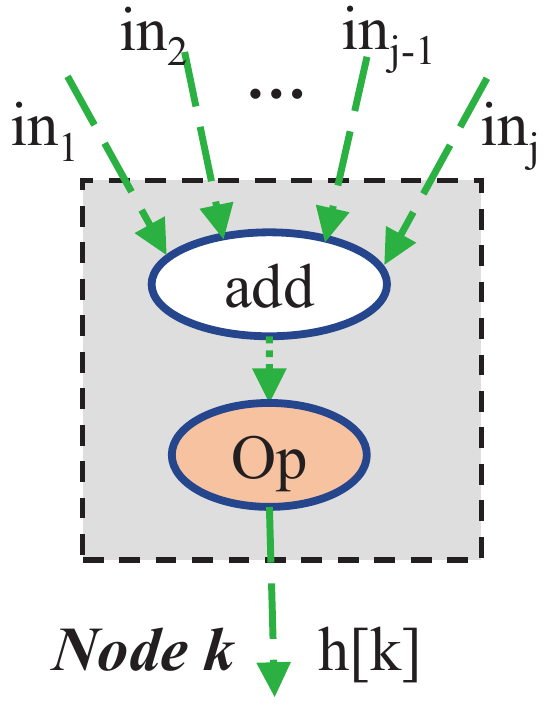}
    }
\caption{The comparison of used nodes in NASNet, one-shot approaches, and our approach.
Here $h[k]$ denotes the output of node $k$ and one of its inputs $in_i$ is selected from $H=\{\widehat{h}[0], \widehat{h}[1], h[0], ..., h[k-1] \}$,
where $\widehat{h}[0]$ and $\widehat{h}[1]$ are the outputs of two previous cells, and the others are the outputs of all precursor nodes.}

\label{Nodes_figure}
\end{center}
\end{figure}

\textbf{Node.}
As shown in Fig.~\ref{search_space} (c), nodes are fundamental components for constructing cells
since links and operation in each node determine the structure of a cell.
There are three types of widely used nodes for existing NAS approaches as shown in Fig.~\ref{Nodes_figure}, namely,
the node in NASNet search space~\cite{zoph2018learning,pham2018efficient,lu2019nsga,luo2018neural},
the node in one-shot NAS approaches~\cite{liu2018darts,brock2018smash,xie2018snas,dong2019one}
and the graph based node~\cite{kandasamy2018neural}.
To be specific,
the node in NASNet (called block) first applies two operations $Op_1$ and $Op_2$ to its two input feature maps $\{in_1, in_2\}$ and
then merges two obtained outputs via element-wise addition to get the final output.
The node in one-shot NAS approaches represents a specific tensor and each edge denotes an operation,
and all previous outputs $H=\{\widehat{h}[0], \widehat{h}[1], h[0], ..., h[k-1] \}$ are passed to the node $k$ via the operations on edges,
where the final output $h[k]$ is obtained by a weighted element-wise addition.
Differently,
the graph based node can flexibly receive $j$ ($j \leq k+2 $) previous outputs $\{in_1,in_2, ..., in_j\}$ from $H$.
After combining the received outputs via an element-wise addition,
an operation $Op$ is applied to achieve its final output $h[k]$.

Particularly, the proposed MFENAS adopts the graph based node considering its superiority in constructing neural architectures over the other two types of nodes.
Compared with the node in NASNet, the graph based node is able to construct more flexible and diverse  neural architectures.
Compared with the node in one-shot approaches, the graph based node can construct neural architectures that are easier to extend,
since adding an extra graph based node brings much less additional operations and parameters than adding an extra node in one-shot approaches.

%

\begin{table}[t!]

\caption{Predefined Operation Search Space As Used In~\cite{luo2018neural,zoph2018learning} and Their Encoding.}
\label{search_space_t}
  \centering
  \begin{tabular}{ccc}
    \toprule
    \textbf{Operation} & \textbf{Short name} & \textbf{Number} \\\hline
    Identity mapping& Identity & 0 \\\hline
    Convolution with kernel size 1*1& Conv 1*1 & 1 \\\hline
    Convolution with kernel size 3*3& Conv 3*3 & 2 \\\hline
    Convolution with kernel size 1*3& \multirow{2}{*} {Conv 1*3+3*1 }& \multirow{2}{*}{3} \\
     and kernel size 3*1&  &  \\\hline
     Convolution with kernel size 1*7& \multirow{2}{*} {Conv 1*7+7*1 }& \multirow{2}{*}{4} \\
     and kernel size 7*1&  &  \\\hline
     Max pooling with kernel size 2*2& MaxPool 2*2 & 5 \\\hline
     Max pooling with kernel size 3*3& MaxPool 3*3 & 6 \\\hline
     Max pooling with kernel size 5*5& MaxPool 5*5 & 7 \\\hline
     Average pooling with kernel size 2*2& AvgPool 2*2 & 8 \\\hline
     Average pooling with kernel size 3*3& AvgPool 3*3 & 9 \\\hline
     Average pooling with kernel size 5*5& AvgPool 5*5 & 10 \\
     \bottomrule
  \end{tabular}
\end{table}

\subsubsection{\textbf{Encoding Strategy}}\indent

In the proposed MFENAS, a neural architecture is encoded by an individual $I=\{\mathbf{NC}, \mathbf{RC}\}$,
where two vectors $\mathbf{NC}$ and $\mathbf{RC}$ represent normal cell and reduction cell, respectively.
Each of the two vectors is composed of some sub-vectors,
where each of these sub-vectors represents a node.
Specifically, the $\mathbf{NC}$ and $\mathbf{RC}$ are denoted as follows:
\begin{equation}
\small
\left\{
\begin{aligned}
\mathbf{NC}=&(\mathbf{n_0},\mathbf{n_1},\cdots,\mathbf{n_k},\cdots,\mathbf{n_{N-1}}), k=0,1,\cdots,\mathbf{N-1}\\
\mathbf{RC}=&(\mathbf{\tilde{n}_0},\mathbf{\tilde{n}_1},\cdots,\mathbf{\tilde{n}_m},\cdots,\mathbf{\tilde{n}_{R-1}}),m=0,1,\cdots,\mathbf{R-1}\\
\end{aligned}
\right.,
\label{encoding_NR}
\end{equation}
where the sub-vector $\mathbf{n_k}$ ($\mathbf{\tilde{n}_m}$) denotes the node $k$ (node $m$) in the normal cell $\mathbf{NC}$ (reduction cell $\mathbf{RC}$),
and $\mathbf{N}$ ($\mathbf{R}$) denotes the number of sub-vectors in $\mathbf{NC}$ ($\mathbf{RC}$).
Each sub-vector consists of an operation in the corresponding node and a set of links for connecting to some precursor nodes of this node.
Specifically, the sub-vector $\mathbf{n_k}$ can be denoted as follows:
\begin{equation}
\small
\left.
\begin{aligned}
\mathbf{n_k} = &(\mathbf{L}, Op)= (\widehat{l}_{0},\widehat{l}_{1},l_0,l_1,\cdots,l_i,\cdots,l_{k-1}, Op),\widehat{l}_0,\widehat{l}_1\in \{0,1\}\\
&i=0,1,\cdots,k-1,\ l_i \in \{0,1\},\ Op \in \{0,1,\cdots,10\}\\
\end{aligned}
\right.,
\label{encoding_node}
\end{equation}
where $\mathbf{L}=(\widehat{l}_{0},\widehat{l}_{1},l_0,l_1,\cdots,l_i,\cdots,l_{k-1})$ is a vector of links in the node $k$,
the $\widehat{l}_{0}$ ($\widehat{l}_{1}$) indicates whether the node $k$ links to the first (second) input node,
the $l_i$ indicates whether the node $k$ links to the node $i$,
and $Op$ denotes the number of a specific operation for the node $k$ according to Table~\ref{search_space_t}.

\begin{figure}[t!]
\begin{center}
\includegraphics[width=0.9\linewidth]{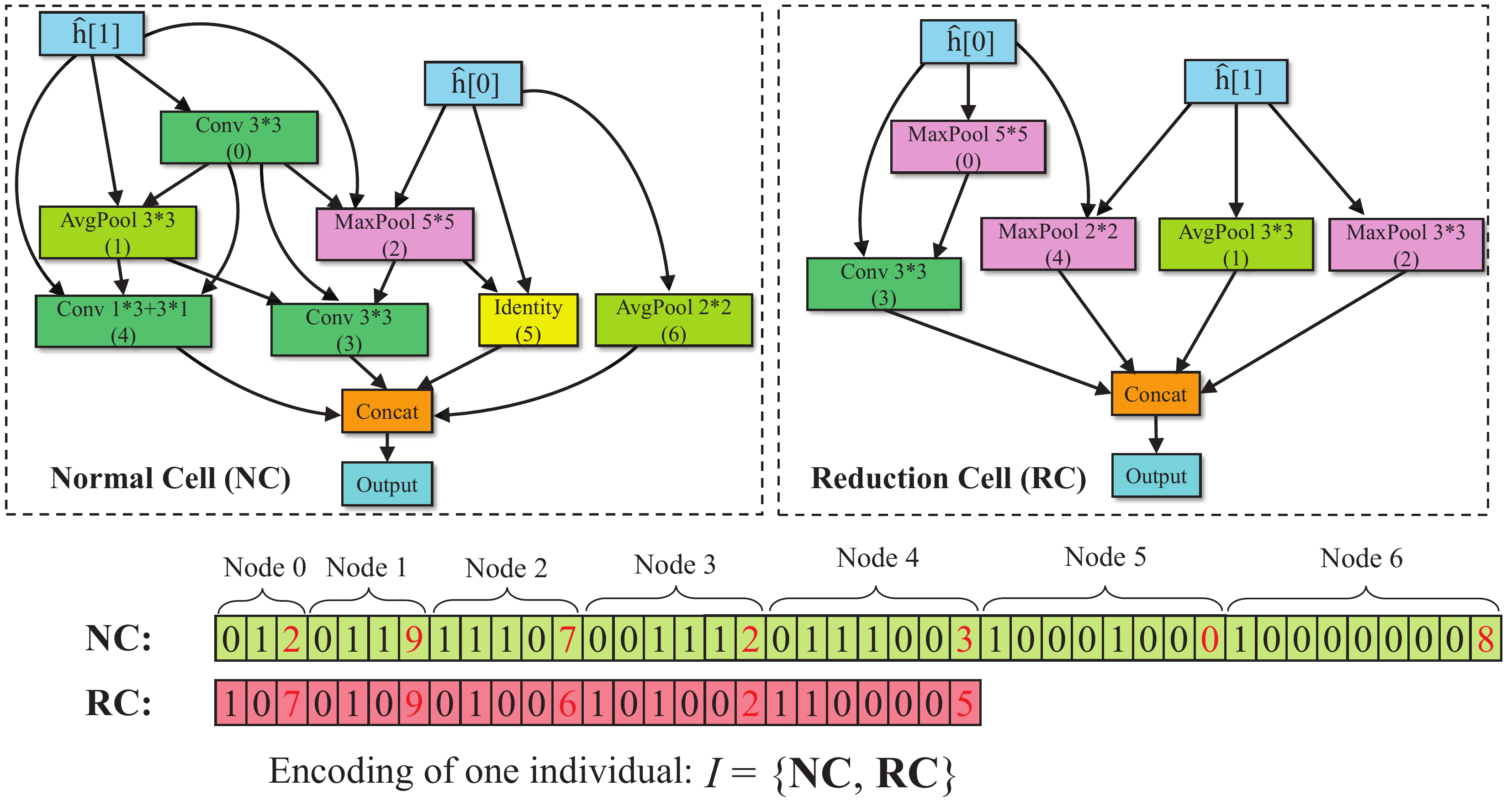}
 \caption{An illustrative example of the proposed encoding strategy.}\label{encoding_fig}
\end{center}
\end{figure}

For better understanding, Fig.~\ref{encoding_fig} gives an illustrative example of the proposed encoding strategy.
As can be seen from Fig.~\ref{encoding_fig}, the encoding $\mathbf{NC}$ with seven sub-vectors represents a normal cell consisting of seven nodes,
where each sub-vector represents a node with specific links and operation.
For example, the node 4 encoded by $(0,1,1,1,0,0,3)$ indicates that the node 4 links to the second input node $\widehat{h}[1]$,
the node 0, the node 1, and adopts the operation Conv 1*3+3*1.

\subsubsection{\textbf{Genetic Operator}}\label{sec:genetic}\indent

Based on the suggested architecture search space, we specially design an effective genetic operator consisting of crossover and mutation for offspring generation in the proposed MFENAS.

\textbf{Crossover.} The crossover operator consists of two components: the inter-cell crossover and the intra-cell crossover.
The inter-cell crossover is to exchange normal cells or reduction cells between two parent individuals.
For example, given two parent individuals $\mathbf{P_1} = \{\mathbf{NC_1},\mathbf{RC_1}\}$
and $\mathbf{P_2} = \{\mathbf{NC_2},\mathbf{RC_2}\}$,
two offspring individuals can be obtained by
\begin{equation}
\small
\left.
\begin{aligned}
 \mathbf{P_1} = \{\mathbf{NC_1},\mathbf{RC_1}\} \\
\mathbf{P_2} = \{\mathbf{NC_2},\mathbf{RC_2}\}\\
\end{aligned}
\right.
\Rightarrow\left.
\begin{aligned}
 \mathbf{Q_1} = \{\mathbf{NC_1},\mathbf{RC_2}\} \\
\mathbf{Q_2} = \{\mathbf{NC_2},\mathbf{RC_1}\}\\
\end{aligned}
\right..
\label{inter_cross}
\end{equation}

The intra-cell crossover is to exchange links and operations in normal cells or reduction cells between two parent individuals.
Specifically, the intra-cell crossover is executed based on a single-point crossover or a revised one-way crossover.
Take the following two normal cells $\mathbf{NC_1}$ and $\mathbf{NC_2}$ as an example:
\begin{equation}
\small
\left\{
\begin{aligned}
\mathbf{NC_1}=&(\mathbf{n_0^1},\mathbf{n_1^1},\cdots,\mathbf{n_k^1}=( \widehat{l}_{0}^{1}, \widehat{l}_{1}^{1},\cdots,\\
&l_i^1,\cdots,l_{k-1}^{1}, Op^{1}),\cdots,\mathbf{n_{N_1-1}^{1}}),\ N_1\ \rm{nodes}\\
\mathbf{NC_2}=&(\mathbf{n_0^2},\mathbf{n_1^2},\cdots,\mathbf{n_k^2}=(\widehat{l}_{0}^{2},\widehat{l}_{1}^{2},\cdots,\\
&l_i^2,\cdots,l_{k-1}^{2}, Op^{2}),\cdots,\mathbf{n_{N_2-1}^{2}}),\  N_2\ \rm{nodes}\\
\end{aligned}
\right.,
\label{example_NC}
\end{equation}
where the length of $\mathbf{NC_1}$ and $\mathbf{NC_2}$ are $\frac{(N_1+5)N_1}{2}$ and $\frac{(N_2+5)N_2}{2}$, respectively.
Suppose $N_1\le N_2$ and a random integer $RI$ is sampled from 0 to $\frac{(N_2+5)N_2}{2}$, then two offspring cells $\mathbf{NC_1^o}$ and $\mathbf{NC_2^o}$ can be generated by a single-point crossover~\cite{tian2019evolutionary}
\begin{equation}
\left.
\small
\begin{aligned}
\rm{when}\ RI \leq \ \frac{(N_1+5)N_1}{2}
\Rightarrow\left\{
\begin{aligned}
&\mathbf{NC_1}[1:RI] \Leftrightarrow \mathbf{NC_2}[1:RI]\\
&\mathbf{NC_1^o}\longleftarrow \ \mathbf{NC_1}\\
&\mathbf{NC_2^o}\longleftarrow \ \mathbf{NC_2}\\
\end{aligned}
\right.
\end{aligned}
\right.
\label{intra_cross_1}
\end{equation}
or a revised one-way crossover
\begin{equation}
\left.
\small
\begin{aligned}
\rm{when}\ \frac{(k+5)k}{2}<  RI \leq \ \frac{(k+6)(k+1)}{2}\\
\Rightarrow
\left\{
\begin{aligned}
\mathbf{n_k^{2 \to 1}}\ =\ &(\mathbf{n_k^{2}}[1:N_1+2],\mathbf{n_k^{2}}[end]),\\
&\ \rm{where} \ N_1-1<k\leq N_2-1&\\
\mathbf{NC_1^o} \longleftarrow &(\mathbf{NC_1}, \mathbf{n_k^{2 \to 1}}) \\
\mathbf{NC_2^o }\longleftarrow &\ \mathbf{NC_2}\\
\end{aligned}
\right.
\end{aligned}
\right.,
\label{intra_cross_2}
\end{equation}
where $\mathbf{NC_1}[1:RI]$ denotes the sub-vector clapped from the first bit to the $RI$-th bit in $\mathbf{NC_1}$,
and $\mathbf{n_k^{2}}[end]$ is the last bit of $\mathbf{n_k^{2}}$.

The $RI \leq \ \frac{(N_1+5)N_1}{2}$ in Equation~(\ref{intra_cross_1}) ensures that the crossover point $RI$ locates in one bit of the shorter parent cell between the two parent cells.
Then $\mathbf{NC_1}[1:RI]$ will be exchanged with $\mathbf{NC_2}[1:RI]$ using a single-point crossover.

The $\frac{(k+5)k}{2}<  RI \leq \ \frac{(k+6)(k+1)}{2}$ in Equation~(\ref{intra_cross_2}) indicates that $RI$ locates in the sub-vector denoting node $k$.
Then a revised one-way crossover is designed to identify useful bits in the node $k$ and copy the identified bits to $\mathbf{NC_1}$.
As presented in Equation~(\ref{intra_cross_2}),  the links $\mathbf{n_k^{2}}[1:N_1+2]$ and the operation $\mathbf{n_k^{2}}[end]$ in the node $k$ are identified as useful bits for $\mathbf{NC_1}$, and thus are copied from $\mathbf{NC_2}$ to $\mathbf{NC_1}$.

\textbf{Mutation.} The mutation operator also consists of two parts: a single-point mutation and an operator of adding an extra node.
Note that the mutation probability for links and that for operations are different in the single-point mutation.
The operator of adding an extra node to $\mathbf{NC_1}$ can be denoted by
\begin{equation}
\small
\left\{
\begin{aligned}
&\mathbf{n_{N1}^{mutation}} \longleftarrow  \rm{Random} (\mathbf{L},Op)\\
&\mathbf{NC_1}\longleftarrow (\mathbf{NC_1}, \mathbf{n_{N1}^{mutation}})\\
\end{aligned}
\right.,
\label{mutation_adding_node}
\end{equation}
where $\mathbf{n_{N1}^{mutation}}$ is a new node generated in a random way,
and is then combined with cell $\mathbf{NC_1}$ into a new $\mathbf{NC_1}$.

\subsubsection{\textbf{Population Initialization Strategy}}\label{section_initialization}\indent

\begin{algorithm}[t!]
\caption{Initialization$(Pop,Node_{range})$}
\label{algorithm_Initialization}
\small

\begin{algorithmic}[1]

\REQUIRE
{$Pop$: Population size;
 $Node_{range}$: Range of the number of nodes in the initialization;

   }
\ENSURE
$\mathbf{P}$: Population;

\STATE $\mathbf{P} \leftarrow \emptyset$;

\FOR{$i=1$ to $Pop$}
\STATE $prob \leftarrow \frac{i}{Pop}$;
\STATE $node_N,node_R \leftarrow$ Generate two random numbers from $Node_{range}$;
\STATE $\mathbf{NC},\mathbf{RC}\leftarrow$ Generate two zero-vectors with length $\frac{(node_N+5)node_N}{2}$ and $\frac{(node_R+5)node_R}{2}$;

\STATE Set the last bit of links in each node to 1 with the probability $1-prob$ for both $\mathbf{NC}$ and $\mathbf{RC}$;
\STATE Set the first two bits of links in each node to 1 with the probability $prob$ for both $\mathbf{NC}$ and $\mathbf{RC}$;
\STATE Randomly set other bits to 1 for both $\mathbf{NC}$ and $\mathbf{RC}$
\STATE Randomly sample one operation from Table~\ref{search_space_t} for each node in $\mathbf{NC}$ and $\mathbf{RC}$;
\STATE Randomly replace operations of $\mathbf{NC}$ by randomly sampled convolutional-like operations(encoding number from 1 to 4);
\STATE Randomly replace operations of $\mathbf{RC}$ by randomly sampled pooling-like operations (encoding number from 5 to 10);
\STATE $\mathbf{P_i} \leftarrow \{\mathbf{NC},\mathbf{RC}\}$;
\STATE $\mathbf{P} \leftarrow \mathbf{P} \cup \mathbf{P_i}$;
\ENDFOR

\RETURN $\mathbf{P}$;
\end{algorithmic}

\end{algorithm}

The population initialization often has a great influence in convergence and diversity of EAs~\cite{tian2019fuzzy,zhang2019indexed}.
Therefore, we design an effective initialization strategy for generating diverse architectures varying from ResNet-like architectures to Inception-like ones.
A ResNet-like architecture is constructed by setting the last bit of links in each node to 1 with a high probability,
while an Inception-like architecture is constructed by setting the first two bits of links in each node to 1 with a high probability.
Hence, it is intuitive to generate diverse architectures by controlling the probability of setting some bits of individuals to 1.

Specifically, Algorithm~\ref{algorithm_Initialization} presents the detailed procedure of
the initialization strategy in the proposed MFENAS, where there is no hyperparameter except for $Node_{range}$ controlling the range of generated models.
First, the probability of setting the first two bits to 1 is determined for initial individuals,
which ranges from $\frac{1}{Pop}$ to $\frac{Pop}{Pop}$ for diversity.
Then, two numbers $node_N$, $node_R$ are randomly generated in a predefined range,
by which the length of $\mathbf{NC}$ and that of $\mathbf{RC}$ are set to $\frac{(node_N+5)node_N}{2}$ and $\frac{(node_R+5)node_R}{2}$, respectively.
Two vectors $\mathbf{NC}$ and $\mathbf{RC}$ are then generated by setting each bit to 0 based on their length.
Next, the last bit of links in each node for both $\mathbf{NC}$ and $\mathbf{RC}$ can be set to 1 with a probability $1-prob$,
whereas the first two bits of links in each node for both $\mathbf{NC}$ and $\mathbf{RC}$ can be set to 1 with a probability $prob$.
The higher the probability $prob$, the closer the decoding of the individual is to the Inception-like architecture.
The lower the probability $prob$, the closer the decoding of the individual is to the ResNet-like architecture.
Other bits of links in each node for both $\mathbf{NC}$ and $\mathbf{RC}$ are randomly set to 1 or 0.
The operations of nodes in $\mathbf{NC}$ and $\mathbf{RC}$ are randomly sampled from number 0 to number 10.
Afterward, operations of $\mathbf{NC}$ and $\mathbf{RC}$ are randomly replaced by sampled operations in convolution operations (from number 1 to number 4)
and pooling operations (from number 5 to number 10), respectively.
The normal cells $\mathbf{NC}$ after the replacement tend to equip with more convolution operations than those without the replacement,
while the reduction cells $\mathbf{RC}$ after the replacement tend to equip with more pooling operations than those without the replacement.
Finally, the generated individual $\mathbf{P_i}$ will be added to population $\mathbf{P}$.


\section{Empirical Studies}\label{sec:exp}
In this section, we first validate the effectiveness of the proposed MFENAS.
Then, we demonstrate the superiority of the proposed MFENAS by comparing it to 22 state-of-the-art NAS approaches.
Finally, we discuss the population initialization and architecture search space that are suggested in the proposed MFENAS.

\subsection{Experiment Settings}
\subsubsection{\textbf{Benchmark Datasets}}\indent

As suggested in most existing ENAS works\cite{zoph2018learning,lu2019nsga,sun2020automatically,pham2018efficient},
we conduct search process of the proposed MFENAS on CIFAR-10 dataset and evaluate the best architecture obtained by the proposed MFENAS on CIFAR-10, CIFAR-100~\cite{krizhevsky2009learning} and ILSVRC 2012 ImageNet~\cite{deng2009imagenet}.

CIFAR-10 is a 10-class natural image dataset consisting of 50,000 training images and 10,000 testing images,
where the size of each color image is $32\times32$.
CIFAR-100 is a dataset which is the same to CIFAR-10 except for that CIFAR-100 has 100 classes.
All images are whitened with the channel mean subtracted and the channel standard deviation divided.
All training images in both search and training process are dealt with the following standard augmentation:
randomly crop $32\times32$ patches from upsampled images of size 40x40 and apply random horizontal flips at a probability of 0.5.
Besides, the cutout augmentation~\cite{devries2017improved} is used for only the training process.
Compared to CIFAR-10 and CIFAR-100, the ImageNet is a more challenging classification dataset
which consists of various resolution images unevenly distributed in 1000 classes.
Furthermore, there are 1.28 million images for the training set and 50,000 images for the validation set in the ImageNet.
According to previous work on NAS, the input image size is set to $224\times224$~\cite{zoph2018learning,luo2018neural,pham2018efficient}.
Besides, we also utilize some commonly used augmentation techniques, i.e., the random resize and crop, the random horizontal flip and the color jitter.

\subsubsection{\textbf{Peer Competitors}}\indent

In order to demonstrate the effectiveness and efficiency of the proposed MFENAS,
various state-of-the-art approaches are selected as peer competitors to compare with MFENAS.
The selected competitors can be roughly divided into three different types~\cite{zhang2020sampled}.
The first type of competitors is the CNN architectures manually designed,
which contains Wide ResNet~\cite{zagoruyko2016wide}, DenseNet~\cite{huang2017densely}, Inception-v1~\cite{szegedy2015going}, MobileNet~\cite{howard2017mobilenets} and ShuffleNet~\cite{zhang2018shufflenet}.
The second type of competitors comprises various non-EA based NAS approaches,
including the reinforcement learning (RL) based approaches: NASNet~\cite{zoph2018learning}, MetaQNN~\cite{baker2016designing}, Block-QNN-S~\cite{zhong2018practical}, EAS~\cite{cai2018efficient}, E-NAS~\cite{pham2018efficient},
and the gradient based approaches: DARTS~\cite{liu2018darts}, NAO~\cite{luo2018neural}.
The third type of competitors refers to ENAS approaches,
including AmoebaNet~\cite{real2019regularized}, Large-scale Evolution~\cite{real2017large}, Hierarchical Evolution~\cite{liu2018hierarchical},
Genetic-CNN~\cite{xie2017genetic}, NSGA-Net~\cite{lu2019nsga}, CNN-GA~\cite{sun2020automatically}, AE-CNN~\cite{sun2019completely}, E2EPP~\cite{sun2019surrogate},
 SI-ENAS~\cite{zhang2020sampled} and CARS-E~\cite{yang2020cars}.

Moreover, we also employ another competitor MFENAS(baseline),
which is the MFENAS without multi-fidelity evaluation.
In MFENAS(baseline), complete epochs are adopted for neural architecture training.

\subsubsection{\textbf{Search Details}}\indent

\textbf{Dataset Details.} According to suggestions in~\cite{zoph2018learning,luo2018neural,sun2020automatically},
the search process is executed on CIFAR-10 dataset, where the original training set is divided into a new training set and validation set (90\%-10\%).

\textbf{Architecture Details.}
As suggested in~\cite{zoph2018learning,lu2019nsga,luo2018neural}, all generated architectures are set to totally hold 5 ($N=1$) cells,
and the number of filters (channels) in each node is set to 16.
The neural architectures will be trained for 25 complete epochs ($S=25$) by the standard SGD optimizer with momentum,
where the learning rate, the momentum rate and the batch size are set to 0.1, 0.9 and 128, respectively.
In addition, a single period cosine decay and L2 weight decay $3\times10^{-4}$ are also utilized.

\textbf{MFENAS Settings.} We adopt the following setting for MFENAS(baseline) and MFENAS:
 population size $Pop=20$, maximum generation $Gen=25$, range of the number of nodes in initialization $Node_{range}=[5,12]$.
The number of fidelities for  multi-fidelity evaluation $MF$ in MFENAS is set to 6 to obtain a better trade-off between the performance and efficiency.
The search process of MFENAS(baseline) and MFENAS is executed on one NVIDIA 2080TI GPU.

\subsubsection{\textbf{Training Details}}\indent

After obtaining Pareto optimal individuals, some promising individuals will be selected to
extend the architectures encoded by the individuals and train them on CIFAR-10, CIFAR-100 and ImageNet.
Here all training settings follow the previous NAS approaches~\cite{zoph2018learning,lu2019nsga,luo2018neural,chen2019renas}.

\textbf{CIFAR-10 and CIFAR-100.} The architecture is set to hold 20 ($N=6$) cells in total,
and the numbers of channels are different in different individuals,
which are set to approximate 3 million architecture parameters based on existing neural architectures~\cite{lu2019nsga}.
When training the obtained architecture for 600 epochs by a standard SGD optimizer with momentum,
the following settings are used: learning rate 0.025, momentum rate 0.9, batch size 128, auxiliary classifier located at $\frac{2}{3}$ of the maximum depth weighted by 0.4, L2 weight decay $5\times10^{-4}$, a single period cosine decay and dropout of 0.4 in the final softmax layer.
Besides, each path will be dropped with a probability 0.2 for regularization introduced in~\cite{zoph2018learning}.

\textbf{ImageNet.} $N$ is set to 4 to make the architecture hold 14 cells,
where the architecture will be trained for 250 epochs by the standard SGD optimizer with the momentum rate that is set to 0.9.
Similarly, a weight decay of $3\times10^{-5}$ and an  auxiliary classifier located at $\frac{2}{3}$ of the maximum depth weighted by 0.4 are also used.
The batch size is set to 512, the learning rate is initially set to 0.1 and later decays by a factor of 0.97 in each epoch.

The training process on both CIFAR-10 and CIFAR-100 is executed on one Tesla P100 GPU,
while the training process on ImageNet is executed two Tesla P100 GPUs.
The source code of proposed MFENAS is available at \url{https://github.com/DevilYangS/MFENAS/}.

\subsection{Effectiveness of Multi-Fidelity Evaluation}\label{sec:tobaseline}

\begin{table}[t!]
  \centering

  \caption{The Performance Comparison of the Best Architecture Found by MFENAS(Baseline) and MFENAS Validated on CIFAR-10, CIFAR-100 and ImageNet Dataset, Respectively. Note that Test Error of ImageNet Refers to Error Rates on Its Validation Dataset.}

       \setlength{\tabcolsep}{0.2mm}{
    \begin{tabular}{ccccc}
    \toprule
    \multicolumn{2}{c}{\multirow{2}[2]{*}{\textbf{Dataset}}} & \textbf{Test } & \textbf{Parameters} & \textbf{Search } \\
    \multicolumn{2}{c}{} & \textbf{Error (\%)} & \textbf{ (M)} & \textbf{Cost} \\
    \midrule
    \multirow{3}[1]{*}{\textbf{MFENAS(baseline)}} & CIFAR-10 & 2.30   & 2.87  & \multirow{3}[1]{*}{2.55} \\
          & CIFAR-100 & 16.18 & 2.89  &  \\
          & ImageNet & 26.16 & 5.79  &  \\
        \midrule
    \multirow{3}[1]{*}{\textbf{MFENAS}} & CIFAR-10 & 2.39  & 2.94   & \multirow{3}[1]{*}{0.6} \\
          & CIFAR-100 & 16.42 & 2.97  &  \\
          & ImageNet & 26.06 & 5.98  &  \\
    \bottomrule
    \end{tabular}}%
     \begin{tablenotes}
        \footnotesize
        \item ``M'' is short for ``million'' in ``Parameters (M)''.
      \end{tablenotes}
  \label{tab:MFENAS(baseline)}%
\end{table}%

\begin{figure}[t!]
\begin{center}
\includegraphics[width=0.7\linewidth]{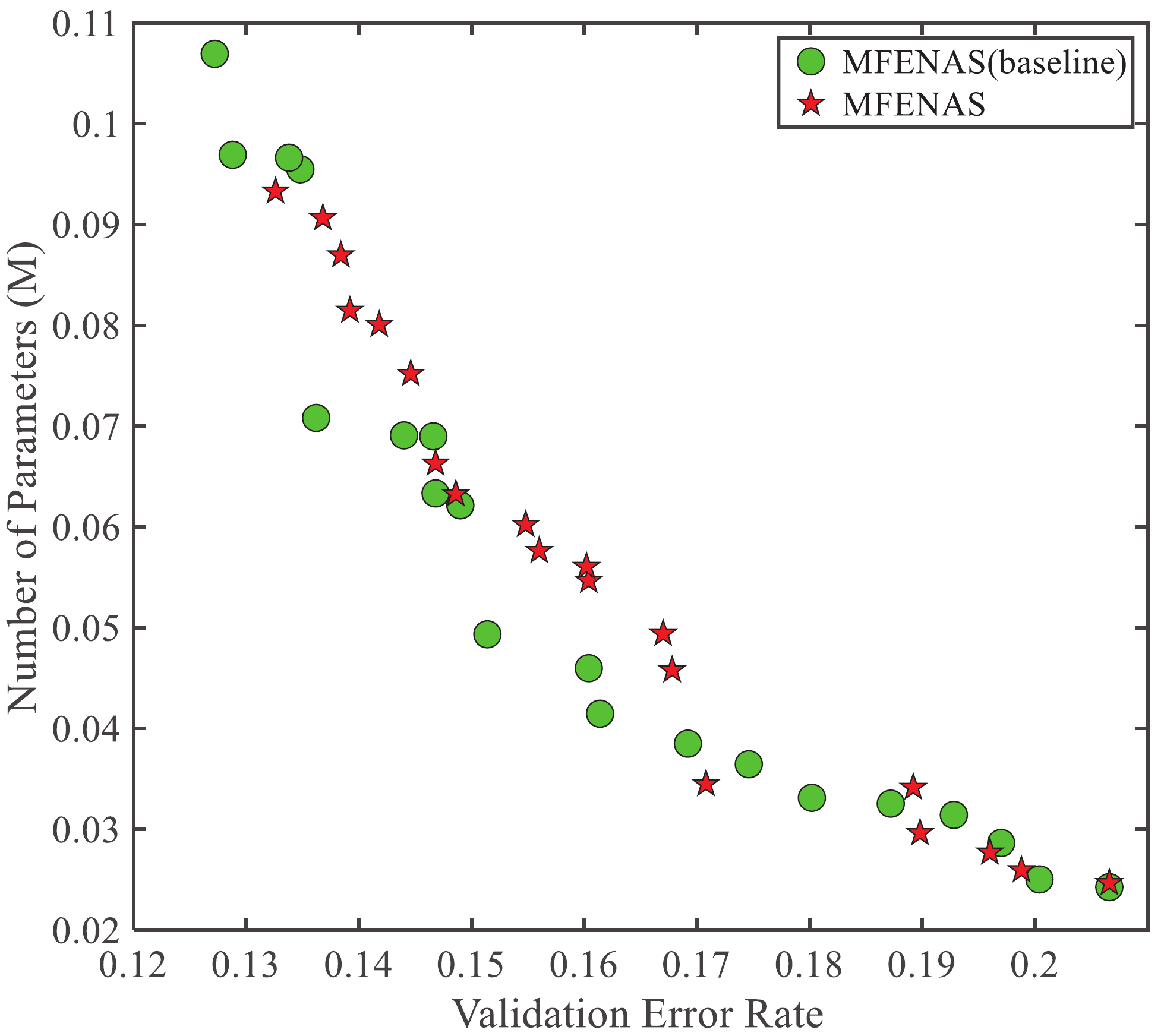}
 \caption{The validation error rate and the number of parameters in the architectures achieved by the proposed MFENAS and MFENAS(baseline).}\label{pareto_optimal_comparison}
\end{center}
\end{figure}

In this section, we will validate the effectiveness of multi-fidelity evaluation in the proposed MFENAS .
First, we verify the performance of the best architecture obtained by MFENAS and MFENAS(baseline) on CIFAR-10, CIFAR-100 and ImageNet.
Table~\ref{tab:MFENAS(baseline)} summarizes comparison results between MFENAS(baseline) and MFENAS
in terms of test error rate, number of model parameters and search cost.
As can be seen from Table~\ref{tab:MFENAS(baseline)}, the search cost of the proposed MFENAS is significantly less than that of MFENAS(baseline), which takes only about $1/4$ search cost of MFENAS(baseline).
Moreover, the architecture found by the proposed MFENAS achieves a similar error rate to that found by MFENAS(baseline) on each of the three datasets.
Particularly, the architecture obtained by the proposed MFENAS has a lower error rate than that obtained by MFENAS(baseline) on the validation dataset ImageNet.
This validates that the multi-fidelity evaluation is capable of accelerating ENAS while maintaining high architecture quality.
To intuitively verify the final architecture quality obtained by the proposed MFENAS,
Fig.~\ref{pareto_optimal_comparison} presents the validation error rate and the number of parameters in the architectures achieved by the proposed MFENAS and MFENAS(baseline).
It can be observed from Fig.~\ref{pareto_optimal_comparison} that the proposed MFENAS finds high-quality architectures that are similar to MFENAS(baseline) in terms of both architecture accuracy and model complexity.

\begin{table*}[t!]
  \centering
  \caption{The Comparison between The Proposed MFENAS and Existing Peer Competitors in Terms of Test Error Rate and Search Cost on The CIFAR-10 Dataset.}
  \setlength{\tabcolsep}{0.7mm}{
    \begin{tabular}{ccccc}
    \toprule
    \multirow{2}[2]{*}{\textbf{Peer Competitors}} & \multirow{2}[2]{*}{\textbf{Test Error (\%)}} & \multirow{2}[2]{*}{\textbf{Parameters (M)}} & \multirow{2}[2]{*}{\textbf{Search Cost (GPU days)}} & \multirow{2}[2]{*}{\textbf{Search Method}} \\
          &       &       &       &  \\
    \midrule
    Wide ResNet & 4.17  & 36.5  & -     & Manual \\
    DenseNet-BC & 3.46  & 25.6  & -     & Manual \\
    \midrule
    NASNet-A & 2.65  & 3.3   & 2000  & RL \\
    Block-QNN-S & 4.38  & 6.1   & 96    & RL \\
    MetaQNN & 6.92  & 11.2  & 80  & RL \\
    EAS   & 4.23  & 23.4  & 10  & RL \\
    E-NAS & 2.89  & 4.6   & 0.5   & RL+weight sharing \\
    DARTS(first order) & 3.00  & 3.3   & 1.5   & Gradient based \\
    DARTS(second order) & 2.76  & 3.3   & 4   & Gradient based \\
    NAO   & 3.18  & 10.6  & 200  & Gradient based \\
    NAO+WS & 3.53  & 2.5   & 0.3   & Gradient based+weight sharing \\
    \midrule
    AmoebaNet-A & 3.34  & 3.2   & 3150  & Evolution \\
    Large-Scale Evolution & 5.40  & 5.4   & 2750  & Evolution \\
    Hierarchical Evolution & 3.63  & 15.7  & 300  & Evolution \\
    Genetic-CNN & 7.10  & -     & 17  & Evolution \\
    NSGA-Net & 2.75  & 3.3   & 4     & Evolution \\
    CNN-GA & 4.78  & 2.9   & 35    &  Evolution \\
    AE-CNN & 4.30  & 2.0  & 27  & Evolution \\
    E2EPP & 5.30  & -     & 8.5   & Evolution \\
    SI-ENAS & 4.07  & -     & 1.8   & Evolution \\
    CARS-E & 2.86  & 3.0   & 0.4   & Evolution+weight sharing \\
    \midrule
    \textbf{MFENAS} & \textbf{2.39 } & 2.94  & 0.6   & Evolution \\
    \bottomrule
    \end{tabular}}%
         \begin{tablenotes}
        \footnotesize
        \item '-' represents that the corresponding result is not publicly available.
      \end{tablenotes}
  \label{tab:cifar10}%
\end{table*}%

\begin{figure}[t!]
\begin{center}
\includegraphics[width=0.7\linewidth]{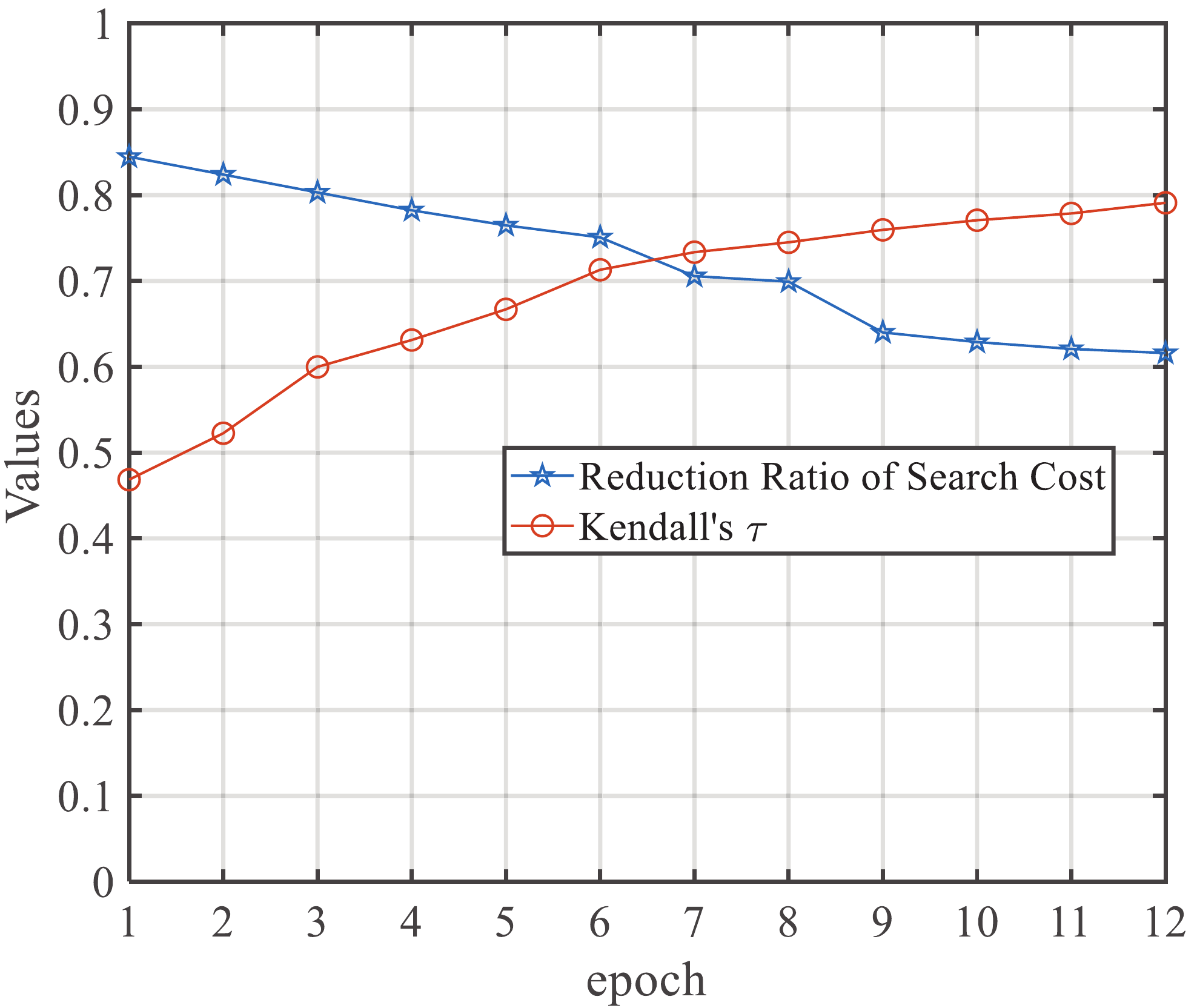}
 \caption{Reduction ratio of search cost and Kendall's $\tau$ with different numbers of fidelities for multi-fidelity evaluation $MF$.}\label{fig:mf}
\end{center}
\end{figure}

For a deeper insight into the multi-fidelity evaluation, we investigate the effect of the number of fidelities $MF$ in the multi-fidelity evaluation
by recording reduction ratio of search cost and Kendall's $\tau$ value at different values of $MF$ ranging from 1 to 12,
which is shown in Fig.~\ref{fig:mf}.
It is seen from Fig.~\ref{fig:mf} that the reduction ratio of search cost decreases with the increase of $MF$,
indicating that the number of fidelities is effective in controlling the search cost of the proposed MFENAS.
The Kendall's $\tau$ value increases with the increase of $MF$,
verifying that the number of fidelities has an influence on the effectiveness of the proposed MFENAS.
Therefore, we set $MF$ to 6 for balancing the efficiency and effectiveness of the proposed MFENAS,
since $MF=6$ is the best trade-off between the reduction ratio of search cost and Kendall's $\tau$ value.
It is also noteworthy that setting $MF$ to 6 leads to over 0.7 Kendall's $\tau$ value,
which is better than the Kendall's $\tau$ value 0.66 achieved by the best known surrogate based ENAS approach~\cite{sun2019surrogate}.
This reveals that the multi-fidelity evaluation is able to balance effectiveness and efficiency of ENAS by tuning the number of fidelities,
which is attributed to the fact that the multi-fidelity evaluation realizes a trade-off between evaluation cost and evaluation accuracy in NAS.

To summarize, the multi-fidelity evaluation is effective in accelerating ENAS and maintaining high architecture quality for solving NAS.

\subsection{Competitiveness of the Proposed MFENAS}

\begin{table}[t!]
  \centering
  \caption{The Comparison between The Proposed MFENAS and Existing Peer Competitors in Terms of Test Error Rate and Search Cost on The CIFAR-100 Dataset. }
  \setlength{\tabcolsep}{0.2mm}{
    \begin{tabular}{ccccc}
    \toprule
    \textbf{Peer} & \textbf{Test} & \textbf{Parameters} & \textbf{Search Cost}  & \textbf{Search} \\
       \textbf{Competitors}   & \textbf{Error (\%)}      &  \textbf{(M)}    &     \textbf{(GPU days)}  & \textbf{Method} \\
    \midrule
    Wide ResNet & 20.50  & 36.5  & -     & Manual \\
    DenseNet-BC & 17.18  & 25.6  & -     & Manual \\
    \midrule
    NASNet-A & 16.58  & 3.3   & 2000  & RL \\
    Block-QNN-S & 20.65 & 6.1   & 96    & RL \\
    MetaQNN & 27.14  & 11.2  & 80  & RL \\
    E-NAS & 17.27  & 4.6   & 0.5   & RL+weight sharing \\
    NAO   & 15.67  & 10.8  & 200  & Gradient based \\
        \midrule
    AmoebaNet-B & 15.80  & 3.2   & 3150  & Evolution \\
    Large-Scale Evolution & 23.00  & 40.4  & 2750  & Evolution \\
    Genetic-CNN & 29.03 & -     & 17  & Evolution \\
    NSGA-Net & 20.74 & 3.3   & 8     & Evolution \\
    CNN-GA & 20.53 & 4.1   & 40    & Evolution \\
    AE-CNN & 20.85  & 5.4  & 36  & Evolution \\
    E2EPP & 22.02  & -     & 8.5   & Evolution \\
    SI-ENAS & 18.64  & -     & 1.8   & Evolution \\
    \midrule
    \textbf{MFENAS} & \textbf{16.42 } & 2.97  & 0.6   & Evolution \\
    \bottomrule
    \end{tabular}}%
     \begin{tablenotes}
        \footnotesize
        \item Note that only 15 peer competitors are adopted for comparison on CIFAR-100 since the architecture quality achieved by the remaining 7 peer competitors on CIFAR-100 is not publicly available.
      \end{tablenotes}
  \label{tab:cifar100}%
\end{table}%

In this section, we discuss the competitiveness of the proposed MFENAS by comparing it with 22 state-of-the-art NAS approaches.
Tables~\ref{tab:cifar10},~\ref{tab:cifar100} and~\ref{tab:imagenet} present the architecture error
rate, the number of parameters and the search cost achieved by the proposed MFENAS and existing peer competitors on CIFAR-10, CIFAR-100 and ImageNet.

As presented in Table~\ref{tab:cifar10}, the architecture found by the proposed MFENAS holds a 2.39\% test error rate on CIFAR-10 dataset,
which is better than test error rates achieved by all the existing peer competitors including manual designed architectures, non-EA based NAS approaches and ENAS approaches.
For the search cost, the proposed MFENAS takes only 0.6 GPU days to achieve the architecture with 2.39\% test error rate on CIFAR-10 dataset,
which is faster than 18 peer competitors but slower than 3 weight sharing-based NAS approaches.
Nevertheless,
the search cost of the proposed MFENAS is very approximate to that of 3 weight sharing-based NAS approaches,
which is only $0.3$ GPU days larger than the fastest weight sharing-based approach.
The proposed MFENAS outperforms 18 peer competitors and is competitive to 3 weight sharing-based NAS approaches in terms of architecture performance and search cost.
Hence, the proposed MFENAS is overall superior over or competitive to state-of-the-art NAS approaches in terms of effectiveness and efficiency on CIFAR-10.

To further investigate competitiveness of the proposed MFENAS,
we transfer the best architecture found by MFENAS from CIFAR-10 with 10 classes to CIFAR-100 with 100 classes.
Table~\ref{tab:cifar100} presents the comparison results between the proposed MFNEAS and existing peer competitors on CIFAR-100.
As can be observed from Table~\ref{tab:cifar100}, the transferred architecture of the proposed MFENAS achieves a 16.42\% test error rate on CIFAR-100,
which performs better than 13 peer competitors but worse than 2 peer competitors (i.e., NAO and AmoebaNet-B).
The reason may lie in the fact that NAO and AmoebaNet-B hold more model parameters than the proposed MFENAS.
However, the search cost of the proposed MFENAS is significantly lower than that of NAO and AmoebaNet-B on CIFAR-100,
indicating that the proposed MFENAS is competitive to NAO and AmoebaNet-B when considering both architecture performance and search cost.
Furthermore, the search cost of the proposed MFENAS is lower than 14 peer competitors but higher than one competitor E-NAS.
The proposed MFENAS still can be regarded as competitive to E-NAS due to its superiority over E-NAS in terms of architecture performance.
Therefore, the proposed MFENAS overall has better or competitive performance compared to state-of-the-art NAS approaches in terms of effectiveness and efficiency on CIFAR-100.

\begin{table}[t!]
\footnotesize
  \centering
  \caption{The Comparison between The Proposed MFENAS and Existing Peer Competitors in Terms of Error Rate and Search Cost on The ImageNet Dataset.}
    \setlength{\tabcolsep}{0.2mm}{
    \begin{tabular}{cccccc}
    \toprule

    \textbf{Peer } & \multicolumn{2}{c}{\textbf{Error (\%)}} & \textbf{Params} & \textbf{Search Cost } & \textbf{Search} \\
\cmidrule{2-3}    \textbf{Competitors} & \textbf{Top-1} & \textbf{Top-5} & \textbf{ (M)} & \textbf{(GPU days)} & \textbf{ Method} \\
    \midrule
    Inception-v1 & 30.20  & 10.10  & 6.6   & -     & Manual \\
    MobileNet & 29.40  & 10.50  & 4.2   & -     & Manual \\
    ShuffleNet (v1) & 29.10  & 10.20  & ~5    & -     & Manual \\
    \midrule
    NASNet-A & 26.00  & 8.4   & 5.3   & 2000  & RL \\
    NASNet-B & 27.2  & 8.7   & 5.3   & 2000  & RL \\
    NASNet-C & 27.5  & 9     & 4.9   & 2000  & RL \\
    DARTS & 26.70  & 8.70  & 4.7   & 4   & Gradient based \\
    NAO   & 25.70  & 8.20  & 11.4  & 200  & Gradient based \\
    AmoebaNet-A & 25.50  & 8.00  & 5.1   & 3150  & Evolution \\
    AmoebaNet-B & 26.00  & 8.50  & 5.3   & 3150  & Evolution \\
    AmoebaNet-C & 24.30  & 7.60  & 6.4   & 3150  & Evolution \\
    Genetic-CNN & 27.87  & 9.74  & 156.0  & 17  & Evolution \\
    \multirow{2}[1]{*}{CARS-E} & \multirow{2}[1]{*}{26.30 } & \multirow{2}[1]{*}{8.40 } & \multirow{2}[1]{*}{4.4 } & \multirow{2}[1]{*}{0.4 } & Evolution+ \\
          &       &       &       &       & weight sharing \\
    \midrule
    \textbf{MFENAS} &   \textbf{26.06 }   &   \textbf{8.18 }   &    \textbf{5.98 }    & \textbf{0.6 } & Evolution \\
    \bottomrule
    \end{tabular}}%
         \begin{tablenotes}
        \footnotesize
        \item Note that the table only compares the peer competitors whose architecture quality is publicly available.
      \end{tablenotes}
  \label{tab:imagenet}%
\end{table}%

In addition to CIFAR-10 and CIFAR-100, we also evaluate the performance of the proposed MFENAS on a more challenging dataset ImageNet,
which holds much more classes, much larger image size and much more images compared to CIFAR-10 and CIFAR-100.
Table~\ref{tab:imagenet} provides the comparison results between the proposed MFENAS and peer competitors on ImageNet,
where the best architecture obtained by the proposed MFENAS is transferred from CIFAR-10 to ImageNet.
It can be seen from Table~\ref{tab:imagenet} that the transferred architecture of the proposed MFENAS holds a 26.06\% error rate,
which is lower than 8 peer competitors but higher than NASNet-A, NAO and AmoebaNet.
Nevertheless, it can still be drawn from Table~\ref{tab:imagenet} that the proposed MFENAS is competitive to NASNet-A, NAO and AmoebaNet on ImageNet, since the search cost of the proposed MFENAS is significantly lower than that of NASNet-A, NAO and AmoebaNet.
More specifically, the search cost of the proposed MFENAS is lower than all the competitors on ImageNet.
Consequently, the performance of the proposed MFENAS overall is better than or competitive to state-of-the-art peer competitors in terms of effectiveness and efficiency on ImageNet.

For summary, the proposed MFENAS is demonstrated to be superior over or competitive to state-of-the-art NAS approaches in terms of effectiveness and efficiency.

\begin{figure}[t!]
\begin{center}
 \centering
    \subfloat[The initial populations obtained by random initialization and the proposed population initialization strategy.]{
        \includegraphics[width=0.45\linewidth]{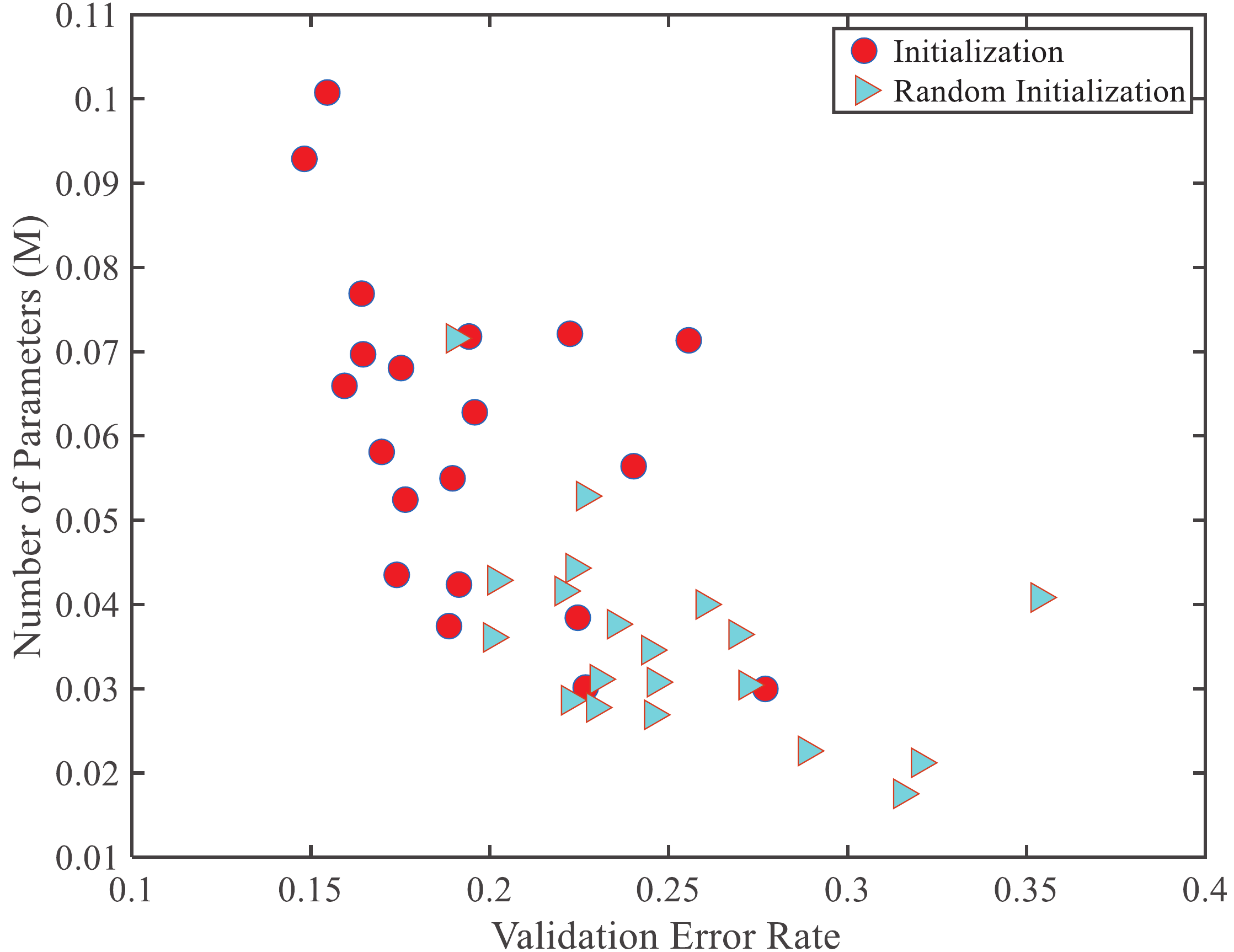}
    }\hfil
    \subfloat[Convergence profiles of HV obtained by random initialization and the proposed population initialization strategy.]{
        \includegraphics[width=0.45\linewidth]{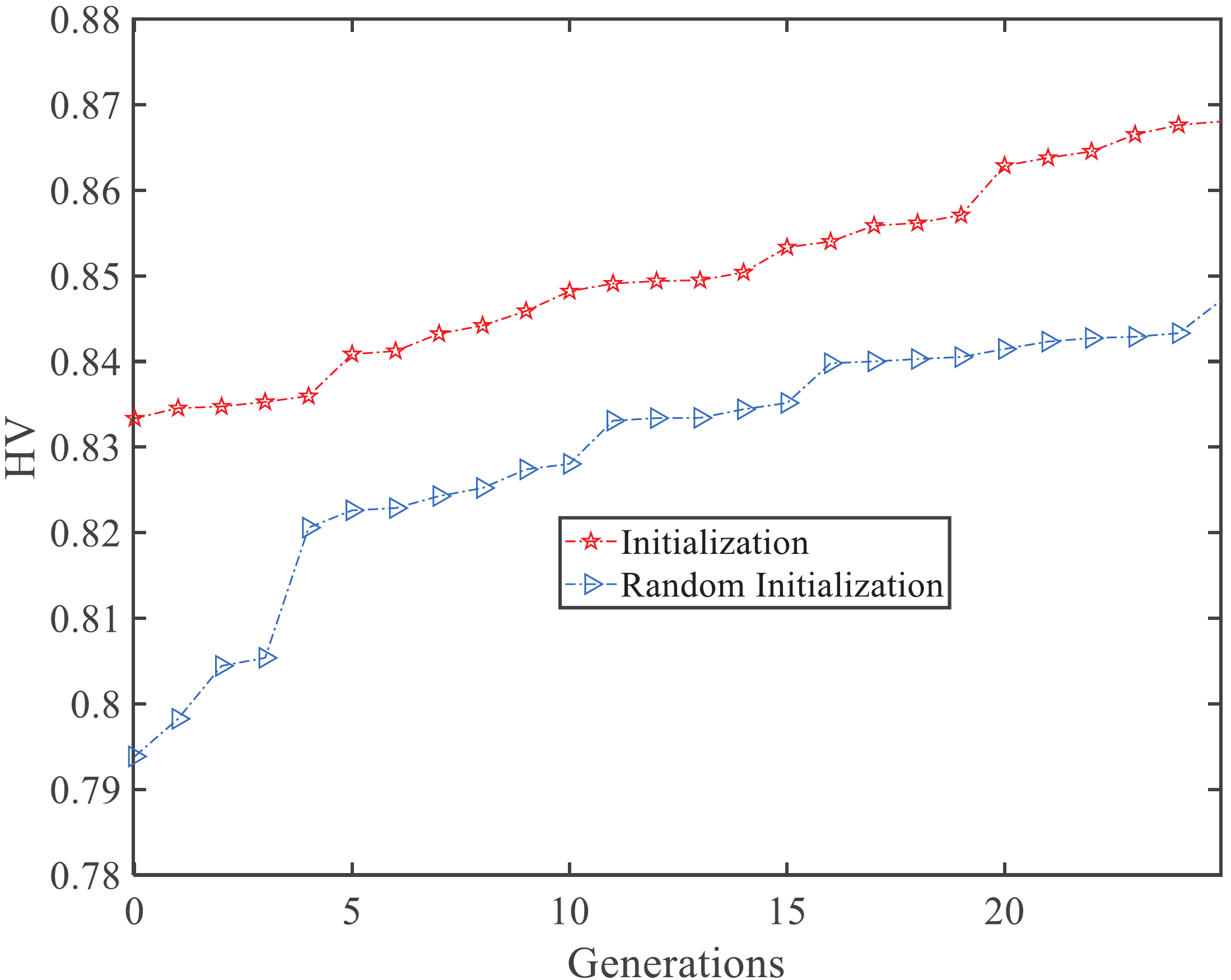}
    }
\end{center}
 \caption{Comparison between random initialization and the proposed population initialization under the MFENAS(baseline) framework.}\label{HV_Initialzition}
\end{figure}

\subsection{Discussions on Details of the Proposed MFENAS}\label{sec:exp_validation}
In this section, we first verify the effectiveness of the population initialization strategy suggested for the proposed MFENAS.
Then we analyze the efficacy of the architecture search space suggested in the proposed MFENAS.

Fig.~\ref{HV_Initialzition} presents the initial populations and the convergence profiles of hypervolume (HV) obtained by random initialization and the proposed population initialization under the MFENAS(baseline) framework,
where HV measures convergence and diversity of a population and a large HV value indicates a good convergence and diversity~\cite{while2006faster}.
Note that the MFENAS(baseline) is adopted as the framework to eliminate the effect of multi-fidelity evaluation in performance of the two initialization strategies.
As can be observed from Fig.~\ref{HV_Initialzition},
the suggested population initialization generates an initial population with better diversity and thus leads to better convergence in MFENAS(baseline) compared to random initialization.
Therefore, the suggested initialization strategy is effective in strengthening performance of the proposed MFENAS by generating diverse architectures in initialization of MFENAS.


\begin{table}[t!]
  \centering
  \caption{Effectiveness Validation of Architecture Search Space by Comparing MFENAS(baseline) and NSGA-Net on CIFAR-10 and CIFAR-100 datasets.}
   \setlength{\tabcolsep}{0.6mm}{
    \begin{tabular}{ccrr}
    \toprule
    \multicolumn{2}{c}{\multirow{2}[2]{*}{\textbf{Dataset}}} & \multicolumn{1}{c}{\textbf{Test }} & \multicolumn{1}{c}{\textbf{Parameters}} \\
    \multicolumn{2}{c}{} & \multicolumn{1}{c}{\textbf{Error (\%)}} & \multicolumn{1}{c}{\textbf{ (M)}} \\
    \midrule
    \multirow{2}[2]{*}{CIFAR-10} & NSGA-Net & 2.75  & 3.3 \\
          & \textbf{MFENAS(baseline)} & \textbf{2.30} & \textbf{2.87} \\
    \midrule
    \multirow{2}[2]{*}{CIFAR-100} & NSGA-Net & 20.74 & 3.3 \\
          & \textbf{MFENAS(baseline)} & \textbf{16.18} & \textbf{2.89} \\
    \bottomrule
    \end{tabular}}%
  \label{tab:baselinevsnsganet}%
\end{table}%

\begin{figure}[t!]
\begin{center}
 \centering
    \subfloat[Normal cell found by MFENAS(baseline).]{\includegraphics[width=0.43\linewidth]{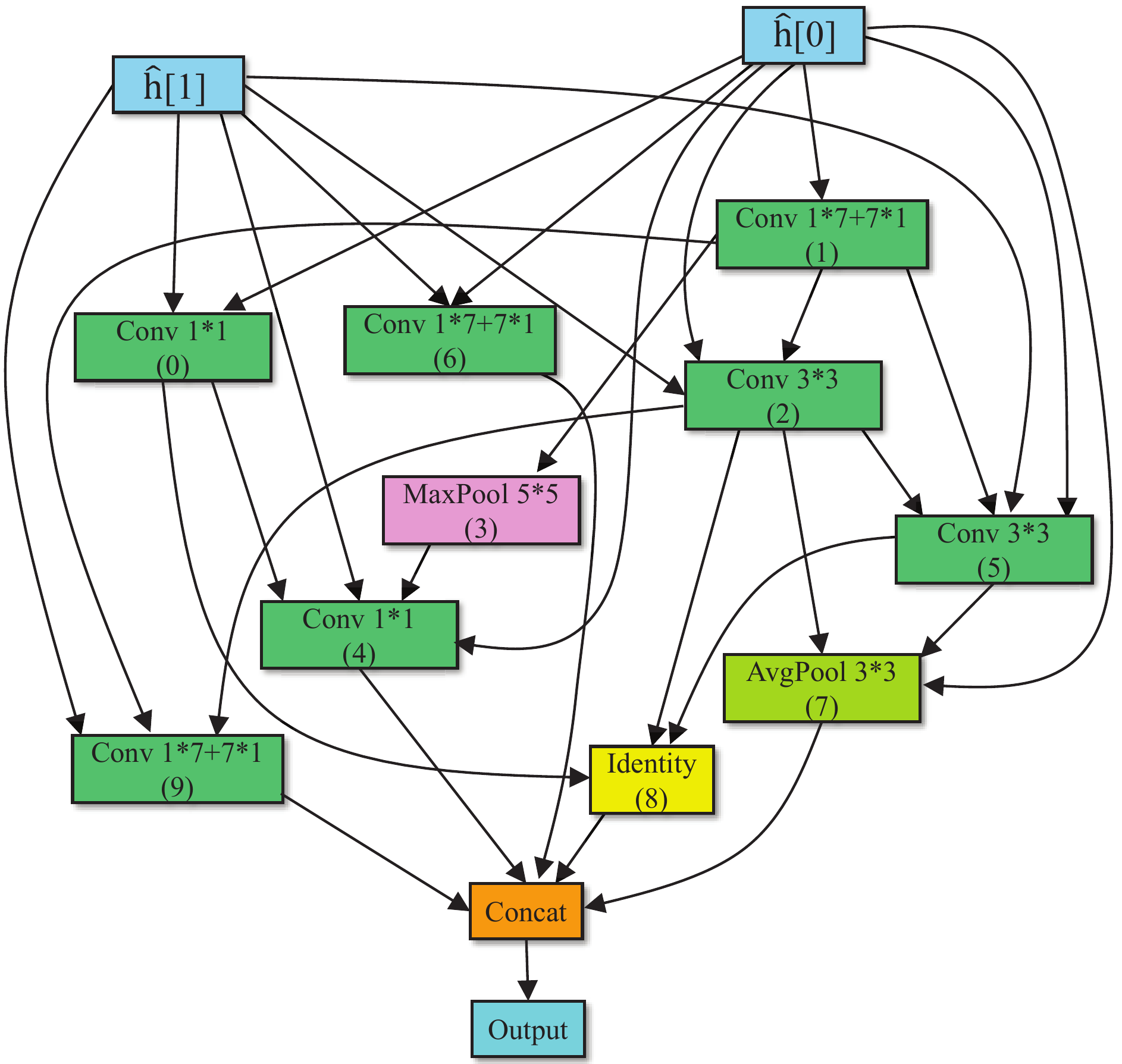}}
    \hfil
    \subfloat[Reduction cell found by MFENAS(baseline)]{\includegraphics[width=0.51\linewidth]{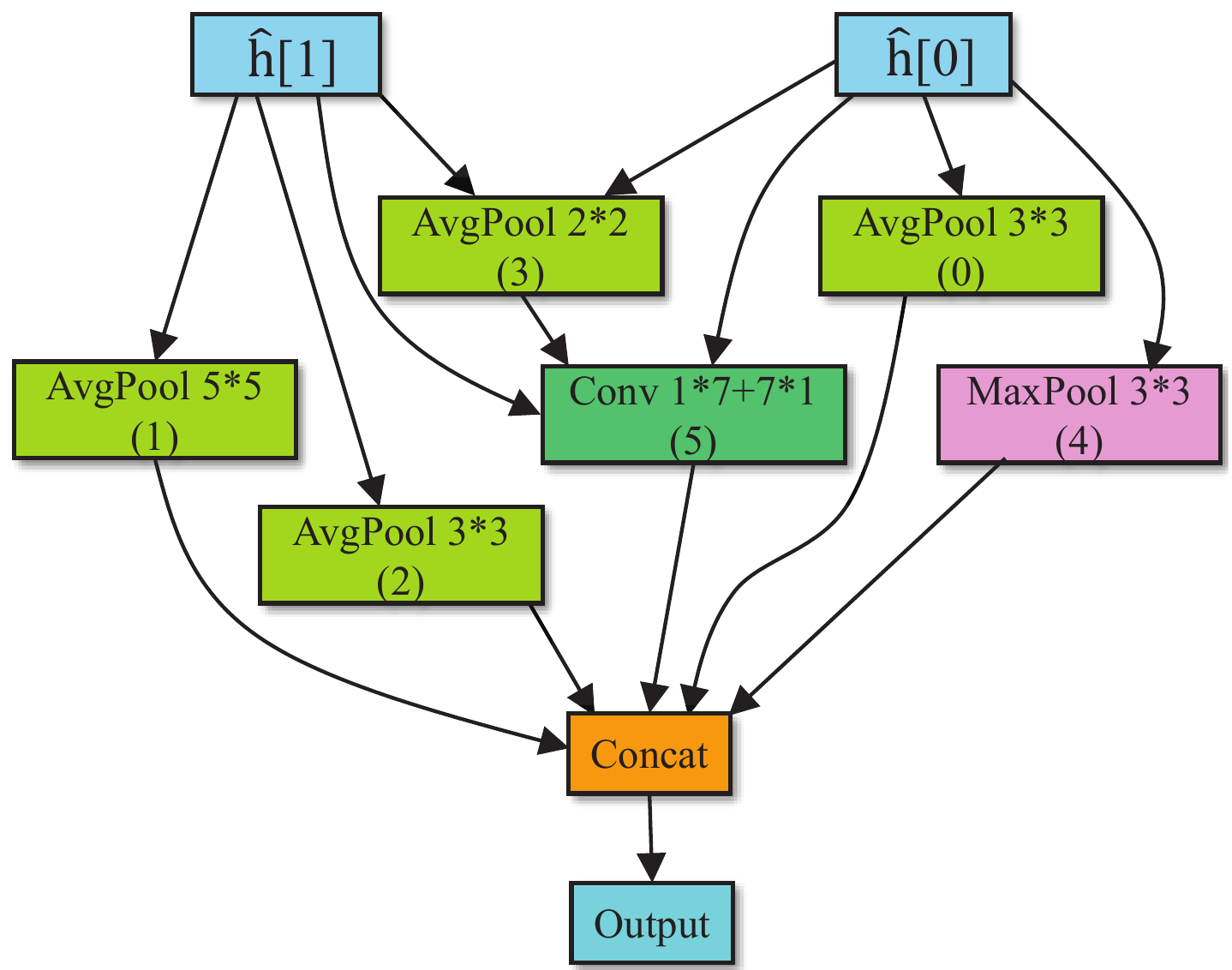}}
    \hfil
    \subfloat[Normal cell found by MFENAS.]{\includegraphics[width=0.47\linewidth]{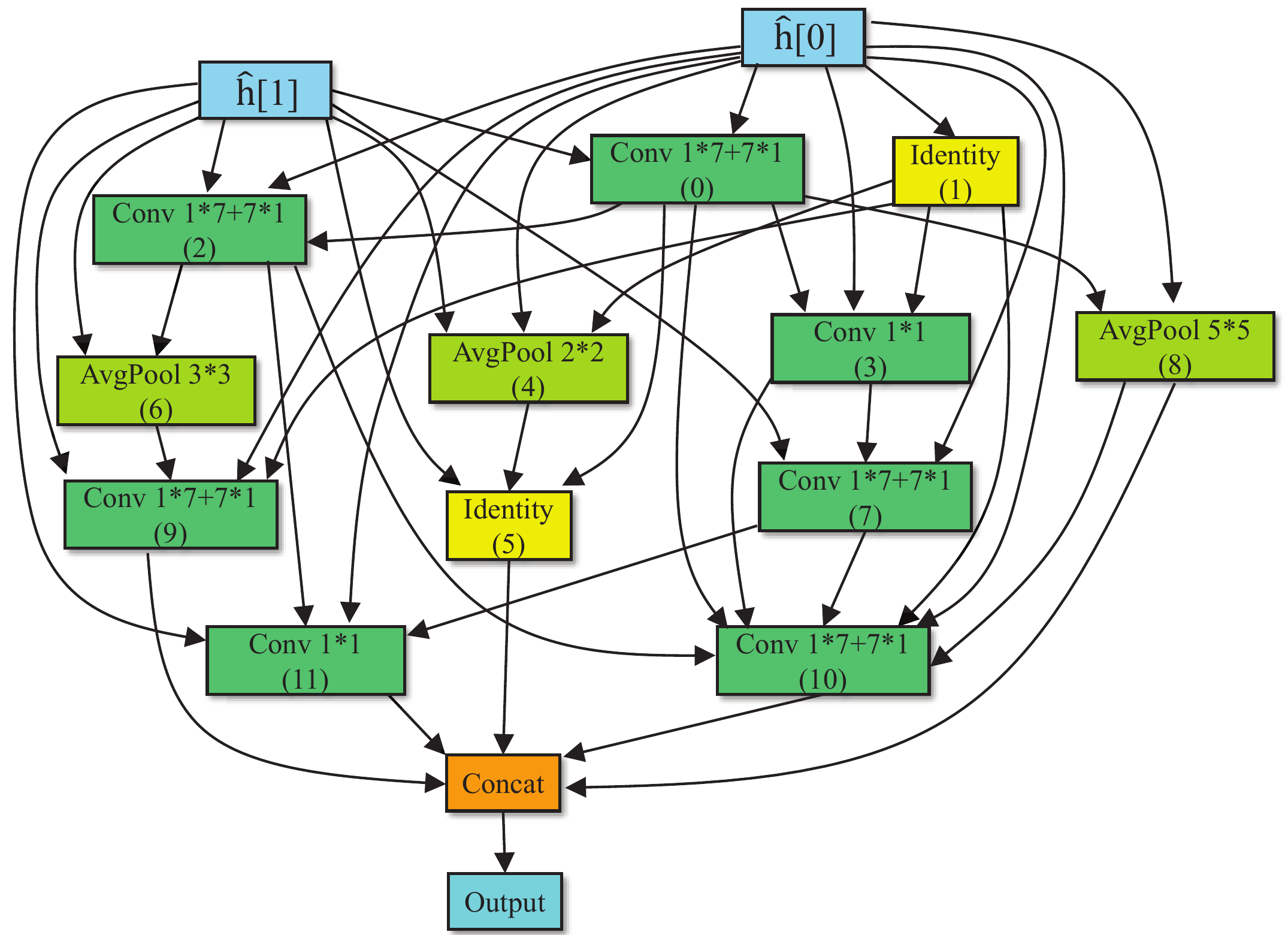}}
    \hfil
    \subfloat[Reduction cell found by MFENAS]{ \includegraphics[width=0.47\linewidth]{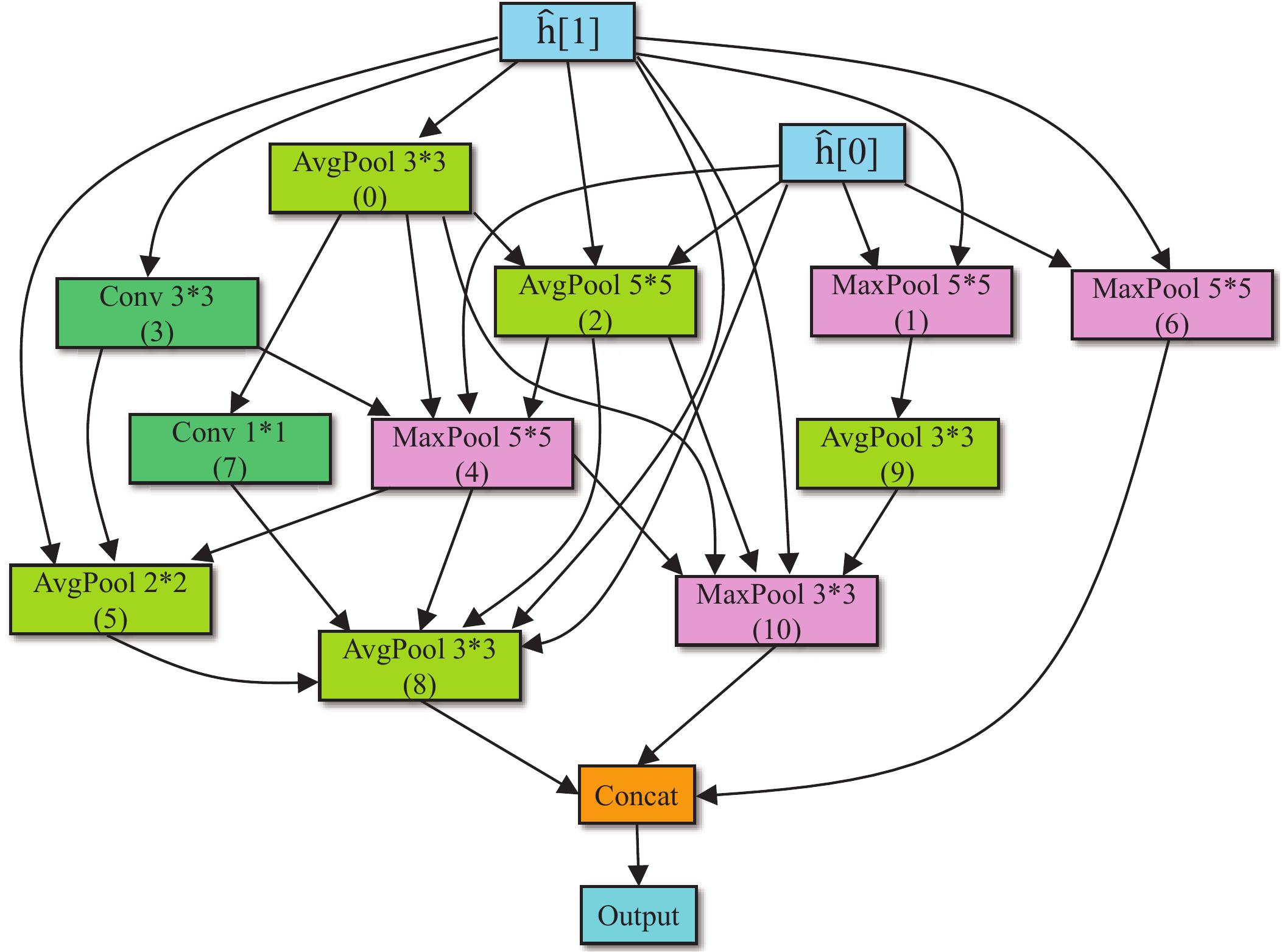}}
\end{center}
\caption{The best architecture obtained by MFENAS(baseline) and MFENAS.}\label{best_architecture}
\end{figure}

To investigate the effect of the suggested architecture search space in the proposed MFENAS,
Table~\ref{tab:baselinevsnsganet} compares the best architecture found by MFENAS(baseline) with that found by NSGA-Net,
which is a fair comparison since the two approaches have the same number of sampled architectures.
MFENAS(baseline) shares the same architecture search space with the proposed MFENAS as suggested in~\ref{sec:sspace},
while NSGA-Net utilizes the NASNet search space.
It can be seen from Table~\ref{tab:baselinevsnsganet} that MFENAS(baseline) can find a higher-quality architecture than NSGA-Net.
Specifically, Fig.~\ref{best_architecture} presents the best architecture obtained by the proposed MFENAS and that obtained by MFENAS(baseline).
The two architectures obtained by the proposed MFENAS and MFENAS(baseline) are quite different from each other under the suggested search space,
while under NASNet search space the best architecture obtained by different NAS approaches are often similar to each other~\cite{elsken2019neural}.
As a result, the suggested search space facilitates NAS approaches in generating more diverse and flexible architectures
than NASNet search space for solving NAS problems.

\section{Conclusions and Future Work}\label{sec:Conclusion}
In this paper, we have developed an accelerated ENAS approach via multi-fidelity evaluation named MFENAS,
to achieve high-quality neural architectures at a low computational cost in NAS.
Empirical results on CIFAR-10 have shown that the architecture found by the proposed MFENAS achieves a 2.39\% test error rate at the cost of only 0.6 GPU days on one NVIDIA 2080TI GPU.
Compared with 22 state-of-the-art NAS approaches, the proposed MFENAS has exhibited superior performance in terms of both computational cost and architecture quality on CIFAR-10.
The architecture transferred to CIFAR-100 and ImageNet has also shown competitive performance to the architectures obtained by existing NAS approaches.

In this paper, the proposed MFENAS mainly concentrates on efficiently finding high-quality architectures,
where the vulnerability of architectures is not considered.
The vulnerability of architectures plays an important role in defensing adversarial attack~\cite{akhtar2018threat}.
Hence, in the future we would like to develop new ENAS approaches to find robust architectures against various adversarial attacks.
Moreover, the binary neural network is characterized with a low memory saving and a low inference latency~\cite{singh2020learning},
but few NAS studies have been reported on searching for binary neural architectures.
Therefore, it is also interesting to design new ENAS approaches for searching high-quality binary neural architectures.

\bibliographystyle{IEEEtran}
\bibliography{mfenas}

\end{document}